\documentclass[sigconf]{acmart}
\usepackage{graphicx}
\usepackage{subcaption}
\usepackage[export]{adjustbox}
\usepackage{booktabs}
\usepackage{multicol, multirow}
\usepackage[normalem]{ulem}
\usepackage{xcolor}
\usepackage{multicol}
\usepackage{enumitem}
\setlist[itemize]{leftmargin=*}
\usepackage{amsmath}
\usepackage{pifont}
\usepackage{algpseudocode}
\usepackage[ruled,linesnumbered]{algorithm2e}
\useunder{\uline}{\ul}{}

\newtheorem{theorem}{Theorem}

\newtheorem{definition}{Definition}

\newcommand\numberthis{\addtocounter{equation}{1}\tag{\theequation}}

\AtBeginDocument{%
  \providecommand\BibTeX{{%
    \normalfont B\kern-0.5em{\scshape i\kern-0.25em b}\kern-0.8em\TeX}}}

\copyrightyear{2024}
\acmYear{2024}
\setcopyright{acmlicensed}\acmConference[WWW '24]{Proceedings of the ACM Web Conference 2024}{May 13--17, 2024}{Singapore, Singapore}
\acmBooktitle{Proceedings of the ACM Web Conference 2024 (WWW '24), May 13--17, 2024, Singapore, Singapore}
\acmDOI{10.1145/3589334.3645703}
\acmISBN{979-8-4007-0171-9/24/05}
\begin{document}

\title{Endowing Pre-trained Graph Models with Provable Fairness}

\author{Zhongjian Zhang}
\authornote{Both authors contributed equally to this research.}
\email{zhangzj@bupt.edu.cn}
\affiliation{
  \institution{Beijing University of Posts and Telecommunications}
  \city{Beijing}
  \country{China}
}

\author{Mengmei Zhang}
\email{zhangmm@bupt.edu.cn}
\authornotemark[1]
\affiliation{
  \institution{Beijing University of Posts and Telecommunications}
  \city{Beijing}
  \country{China}
}

\author{Yue Yu}
\email{yuyue1218@bupt.edu.cn}
\affiliation{
  \institution{Beijing University of Posts and Telecommunications}
  \city{Beijing}
  \country{China}
}

\author{Cheng Yang}
\authornote{Corresponding authors.}
\email{yangcheng@bupt.edu.cn}
\affiliation{
  \institution{Beijing University of Posts and Telecommunications}
  \city{Beijing}
  \country{China}
}

\author{Jiawei Liu}
\email{liu_jiawei@bupt.edu.cn}
\affiliation{
  \institution{Beijing University of Posts and Telecommunications}
  \city{Beijing}
  \country{China}
}

\author{Chuan Shi}
\email{shichuan@bupt.edu.cn}
\authornotemark[2]
\affiliation{
  \institution{Beijing University of Posts and Telecommunications}
  \city{Beijing}
  \country{China}
}

\renewcommand{\shortauthors}{Zhongjian Zhang et al.}

\begin{abstract}
Pre-trained graph models (PGMs) aim to capture transferable inherent structural properties and apply them to different downstream tasks. Similar to pre-trained language models, PGMs also inherit biases from human society, resulting in discriminatory behavior in downstream applications. The debiasing process of existing fair methods is generally coupled with parameter optimization of GNNs. However, different downstream tasks may be associated with different sensitive attributes in reality, directly employing existing methods to improve the fairness of PGMs is inflexible and inefficient. Moreover, most of them lack a theoretical guarantee, i.e., provable lower bounds on the fairness of model predictions, which directly provides assurance in a practical scenario. 
To overcome these limitations, we propose a novel adapter-tuning framework that endows pre-trained \textbf{Graph} models with \textbf{P}rovable f\textbf{A}i\textbf{R}ness (called GraphPAR\footnote{The source code can be found at \href{https://github.com/BUPT-GAMMA/GraphPAR}{https://github.com/BUPT-GAMMA/GraphPAR}.}). GraphPAR freezes the parameters of PGMs and trains a parameter-efficient adapter to flexibly improve the fairness of PGMs in downstream tasks. Specifically, we design a sensitive semantic augmenter on node representations, to extend the node representations with different sensitive attribute semantics for each node. The extended representations will be used to further train an adapter, to prevent the propagation of sensitive attribute semantics from PGMs to task predictions. Furthermore, with GraphPAR, we quantify whether the fairness of each node is provable, i.e., predictions are always fair within a certain range of sensitive attribute semantics. 
Experimental evaluations on real-world datasets demonstrate that GraphPAR achieves state-of-the-art prediction performance and fairness on node classification task. Furthermore, based on our GraphPAR, around 90\% nodes have provable fairness.

\end{abstract}



\maketitle
\section{Introduction}
Graph Neural Networks (GNNs)~\cite{wu2020comprehensive, wang2017community} have achieved significant success in analyzing graph-structured data, such as social networks~\cite{guo2020deep} and webpage network ~\cite{wu2022semi}. Recently, inspired by pre-trained language models, various pre-trained graph models (PGMs)~\cite{xia2022simgrace, DGI2018,hu2020pretraining} have been proposed. Generally, PGMs capture transferable inherent graph structure properties through unsupervised learning paradigms in the pre-training phase, and then adapt to different downstream tasks by fine-tuning. As a powerful learning paradigm, PGMs have received considerable attention in the field of graph machine learning and have been broadly applied in various domains, such as recommendation systems~\cite{hao2021pre} and drug discovery~\cite{wang2102molclr}.

\begin{figure}[tbp]
    \centering
    \begin{subfigure}[b]{0.235\textwidth}
    \includegraphics[width=\linewidth]{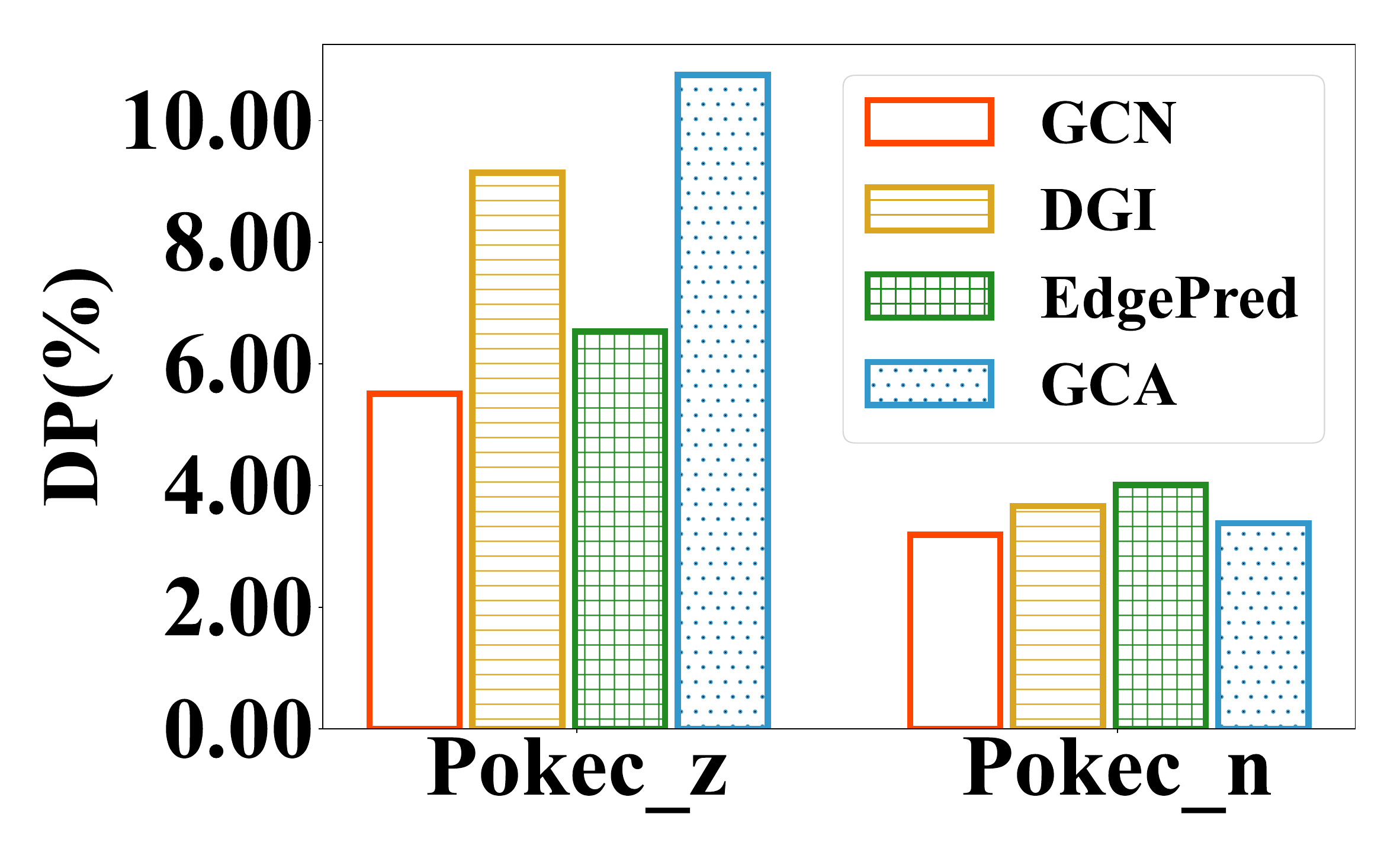}
    \vspace{-15pt}
    \caption{Demographic Parity (DP).}
    \end{subfigure}
    \begin{subfigure}[b]{0.235\textwidth}
    \includegraphics[width=\linewidth]{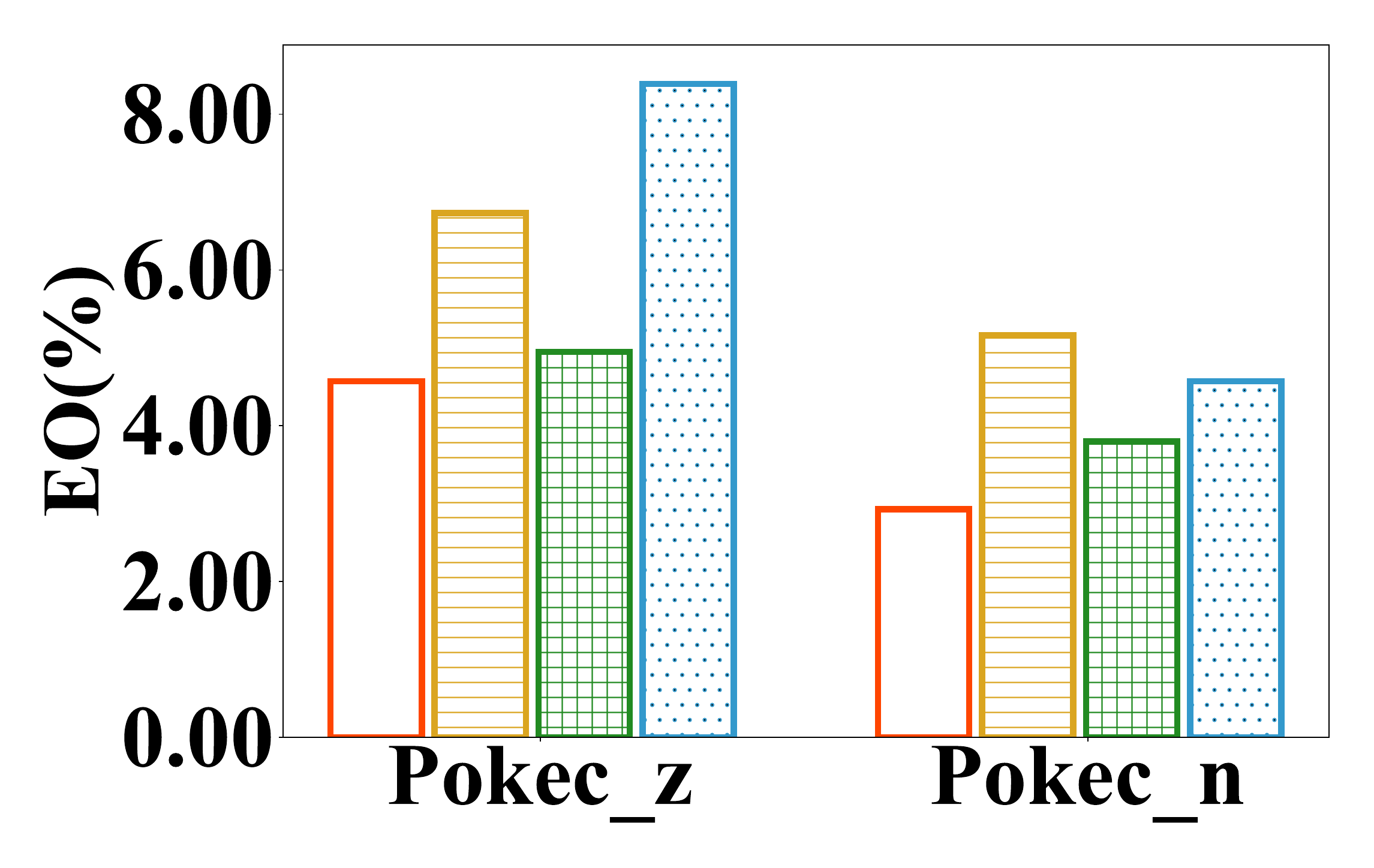}
    \vspace{-15pt}
    \caption{Equality Opportunity (EO).}
    \end{subfigure}
    \vspace{-17pt}
    \caption{An example of evaluating the node classification fairness of PGMs on Pokec\_z and Pokec\_n datasets. DP (↓) and EO (↓) report the fairness of three PGMs (i.e., DGI, EdgePred, GCA) and vanilla GCN.}
    \label{fig:pgms_bias}
    \vspace{-20pt}
\end{figure}

However, recent works~\cite{meade2022empirical, gira2022debiasing} have demonstrated that pre-trained language models tend to inherit bias from pre-training corpora, which may result in biased or unfair predictions towards sensitive attributions, such as gender, race and religion. With the same paradigm, PGMs raise the following question: \textit{Do pre-trained graph models also inherit bias from graphs?} To answer this question, we evaluate the node classification fairness of three different PGMs on datasets Pokec\_z and Pokec\_n~\cite{takac2012data}, the results as depicted in Figure~\ref{fig:pgms_bias}. We observe that PGMs are more unfair than vanilla GCN. This is because PGMs can well capture semantic information on graphs during the pre-training phase, which inevitably contains sensitive attribute semantics. A further question naturally arises: \textit{How to improve the fairness of PGMs?} Addressing this problem is highly critical, especially in graph-based high-stake decision-making scenarios, such as social networks~\cite{khajehnejad2020adversarial} and candidate-job matching~\cite{kenthapadi2017personalized}, because biased predictions will raise severe ethical and societal issues.

Although substantial methods have been proposed for developing fair GNNs in recent years, directly employing them to improve the fairness of PGMs is inflexible and inefficient. Namely, most existing works generally train a fair GNN for a specific task. For example, methods based on counterfactual fairness ~\cite{agarwal2021towards, guo2023improving, ma2022learning} train a fair GNN encoder by constructing different counterfactual training samples. Approaches with sensitive attribute classifiers ~\cite{dai2021say, wang2022improving, ling2022learning} constrain GNNs to capture sensitive semantic information by introducing an adversarial loss in the training phase. Obviously, in the above methods, the debiasing process is coupled with the parameter optimization of GNNs. However, in reality, the same PGM can be used in different downstream tasks and different downstream tasks may be associated with different sensitive attributes~\cite{creager2019flexibly, wu2022selective, Oneto2020FairnessIM}. Debiasing for a specific task in the pre-training phase is inflexible, and maintaining a specific PGM for each task is inefficient. Besides, most existing fairness methods lack theoretical analysis and guarantees ~\cite{chen2023fairness, jovanovic2023fare}, meaning that they do not provide a practical guarantee, i.e., provable lower bounds on the fairness of model prediction. This is significant for determining whether to deploy models in practical scenarios ~\cite{jillson2021aiming, ruoss2020learning, com2021laying, peychev2022latent}.

In this paper, we attempt to address the above questions by proposing GraphPAR, a novel adapter-tuning framework for efficiently and flexibly endowing PGMs with provable fairness.
Specifically, in downstream tasks, we first freeze the parameters of PGMs, then design a sensitive semantic augmenter on node representations, to extend the node representations with different sensitive attribute semantics for each node. 
The extended representations will be directly used to tune an adapter so that the adapter-processed node representations are independent of sensitive attribute semantics, preventing the propagation of sensitive attribute semantics from PGMs to task predictions. 
Furthermore, with GraphPAR, we quantify whether the fairness of each node is provable, i.e., predictions are always fair within a certain range of sensitive attribute semantics. For example, when a person’s gender semantics gradually transit from male to female, our provable fairness guarantees that the prediction results will not change. In summary, GraphPAR can apply to any PGMs while providing fairness with theoretical guarantees. 

Our main contributions can be summarized as follows:
\begin{itemize}
\item We first explore the fairness of PGMs and find that PGMs can capture sensitive attribute semantics during the pre-training phase, leading to unfairness in downstream task predictions. 
\item We propose GraphPAR for efficiently and flexibly endowing PGMs with provable fairness. Specifically, during the adaption of downstream tasks, GraphPAR utilizes an adapter for parameter-efficient adaption and a sensitive semantic augmenter for fairness with practical guarantees.
\item Extensive experiments on different real-world datasets demonstrate that GraphPAR achieves state-of-the-art prediction performance and fairness. Moreover, with the help of GraphPAR, around 90\% of nodes have provable fairness.
\end{itemize}
\section{Related Work}
\subsection{Pre-trained Graph Models}  
Inspired by pre-trained language models, pre-trained graph models (PGMs) capture transferable inherent graph structure properties through unsupervised learning paradigms during the pre-training phase, and then adapt to different downstream tasks by fine-tuning. Based on different pre-training methods, the existing PGMs can mainly be categorized into contrastive and predictive pre-training. Contrastive pre-training maximizes mutual information between different views, encouraging models to capture invariant semantic information across various perspectives. For example, DGI ~\cite{DGI2018} and InfoGraph ~\cite{sun2019infograph} generate expressive representations for nodes or graphs by maximizing the mutual information between graph-level and substructure-level. GraphCL ~\cite{you2020graph} and its variants ~\cite{you2021graph, sun2021mocl} further introduce a range of sophisticated augmentation strategies for constructing different views. Unlike contrastive pre-training, predictive pre-training enables models to understand the universal structural and attribute semantics of graphs. For instance, attribute masking is proposed by ~\cite{hu2020pretraining} where the input node attributes or edge are randomly masked, and the GNN is asked to predict them. EdgePred ~\cite{hamilton2017inductive} samples negative edges and trains a general GNN to predict edge existence. GraphMAE ~\cite{hou2022graphmae} incorporates feature reconstruction and a re-mask decoding strategy to pre-train a GNN.

Despite the ability of PGMs to capture abundant knowledge that proves valuable for downstream tasks, the conventional fine-tuning process still has some drawbacks, such as overfitting, catastrophic forgetting, and parameter inefficiency\cite{li2023adaptergnn}. To alleviate these issues, recent research has focused on developing parameter-efficient tuning (delta tuning) techniques that can effectively adapt pre-trained models to downstream tasks ~\cite{ding2022delta}. Delta tuning ~\cite{ding2022delta} seeks to tune a small portion of parameters and keep the left parameters frozen. For example, prompt tuning ~\cite{liu2023graphprompt} aims to modify model inputs rather than PGMs parameters. Adapter tuning ~\cite{li2023adaptergnn} trains only a small fraction of the adapter parameters to adapt pre-trained models to downstream tasks. 

Though a large number of researches have been proposed on how to design pre-training methods and fine-tune PGMs in downstream tasks, most of them focus on improving performance while ignoring their plausibility in fairness and so on.

\subsection{Fairness of Graph}
Recent study ~\cite{dai2021say} shows that GNNs tend to inherit bias from training data and the message-passing mechanism of GNNs could magnify the bias. Hence, many efforts have been made to develop fair GNNs. According to the stage at which the debiasing process occurs, the existing methods could be split into the pre-processing, in-processing, and post-processing methods~\cite{chen2023fairness}. Pre-processing methods remove bias before GNN training occurs by targeting the input graph structure, input features, or both. For instance, EDITS ~\cite{dong2022edits} propose a novel approach that utilizes the Wasserstein distance to mitigate both attribute and structural bias on graphs. 
In-processing methods focus on modifying the objective function of GNNs to learn fair or unbiased embeddings during training. For example, NIFTY ~\cite{agarwal2021towards} proposes a novel multiple-objective function incorporating fairness and stability. Graphair ~\cite{ling2022learning} introduces an automated augmentation model that generates a graph for fairness and informativeness. 
A few post-processing methods have been proposed to remove bias from GNNs. FairGNN ~\cite{dai2021say} trains a fair GNN encoder through an adversary task of predicting sensitive attributes. FLIP ~\cite{masrour2020bursting} further addresses the problem of link prediction homophily by adversarial learning. 

Although all the methods above have achieved significant success in graph fairness, most of them require optimizing the parameters of GNN. Since different downstream tasks may be associated with different sensitive attributes, these methods cannot flexibly and efficiently improve the fairness of PGMs. Besides, they all lack theoretical analysis and fairness guarantees, which are significant for determining whether to deploy a model in a real-world scenario.
\section{Preliminary}

\subsection{Notations}
Given an attributed graph as $\mathcal G=(\mathcal{V},\mathcal{E},\mathbf{X})$, where $\mathcal{V} = \{v_1,...,v_n\}$ represents the set of $n$ nodes, $\mathcal{E}\subseteq\mathcal{V}\times\mathcal{V}$ represents the set of edges, $\mathbf{X}=\{\mathbf{x}_{1},\dots,\mathbf{x}_{n}\}$ represents the node features and $\mathbf{x}_i \in \mathbb{R}^d$. The adjacency matrix of the graph $\mathcal G$ is denoted as $\mathbf{A}\in\mathbb{R}^{n\times n}$, where $\mathbf{A}_{ij}=1$ if nodes $v_i$ and $v_j$ are connected, otherwise $\mathbf{A}_{ij}=0$. Each node $i$ is associated with a binary sensitive attribute $s_i \in \{0, 1\}$ (we assume one single, binary sensitive attribute for simplicity, but our method can easily handle multivariate sensitive attributes as well). Furthermore, we consider a PGM or GNN denoted as $f$, which takes the graph structure and node features as input and produces node representations. The encoded representations for the $n$ nodes are denoted by $\mathbf{H}=\{\mathbf{h}_i\}_{i=1}^{n}$, where $\mathbf H=f(\mathcal{V},\mathcal{E},\mathbf{X})$ and $\mathbf{h}_i \in \mathbb{R}^p$. 

In the pre-training phase, the parameters of a PGM $f$ are optimized via self-supervised learning, such as graph contrastive learning ~\cite{you2021graph, you2020graph, sun2019infograph, DGI2018} or graph context prediction~\cite{hou2023graphmae2,hou2022graphmae,hamilton2017inductive,hu2020pretraining}. In the downstream adaption phase, adapter tuning freezes the parameter $f_{\theta}$ of the PGM $f$ and tunes parameter $g_{\theta}$ of an adapter $g$ to adapt PGM for different downstream tasks. Generally, $\lvert g_{\theta} \rvert \ll \lvert f_{\theta} \rvert$, where $\lvert \cdot \rvert$ denotes the number of parameters. In the adapter, given an input $\mathbf{h}_i \in \mathbb{R}^{p}$, a down projection projects the input to a $q$-dimensional space, after which a nonlinear function is applied. Then the up-projection maps the $q$-dimensional representation back to $p$-dimensional space. 

\subsection{Fairness Definition on Graph}
The fairness definition on the graph refers to the model prediction results not being influenced by sensitive attributes of nodes, i.e., the prediction results will not change as the sensitive attribute value variations~\cite{ma2022learning}. 
\begin{definition}[Fairness on Graph]\label{def: graph_count_fair}
Given a graph  $\mathcal G=(\mathcal{V},\mathcal{E},\mathbf{X})$, the encoder $f(\cdot)$ and the classifier $d(\cdot)$ trained on this graph satisfies fairness if for any node $v_i$:
\begin{equation}
P( (\hat{y}_i)_{S\leftarrow{s}} | \mathbf{X},\mathbf{A}) = P((\hat{y}_i))_{S\leftarrow{s}^{\prime}}|\mathbf{X},\mathbf{A}), \quad \text{ s.t. } \ \forall s\neq{s}^{\prime},
\end{equation}
where $\hat{y}_i= d \circ f(\mathbf{X}, \mathbf{A})_i$ denotes the predicted label for node $v_i$, and $s, s^{\prime} \in \{0, 1\}^n$ are two arbitrary sensitive attribute values. 
\end{definition}
 
\section{The Proposed Framework}

\begin{figure*}[!htbp]
    \centering
    \includegraphics[width=1\linewidth]{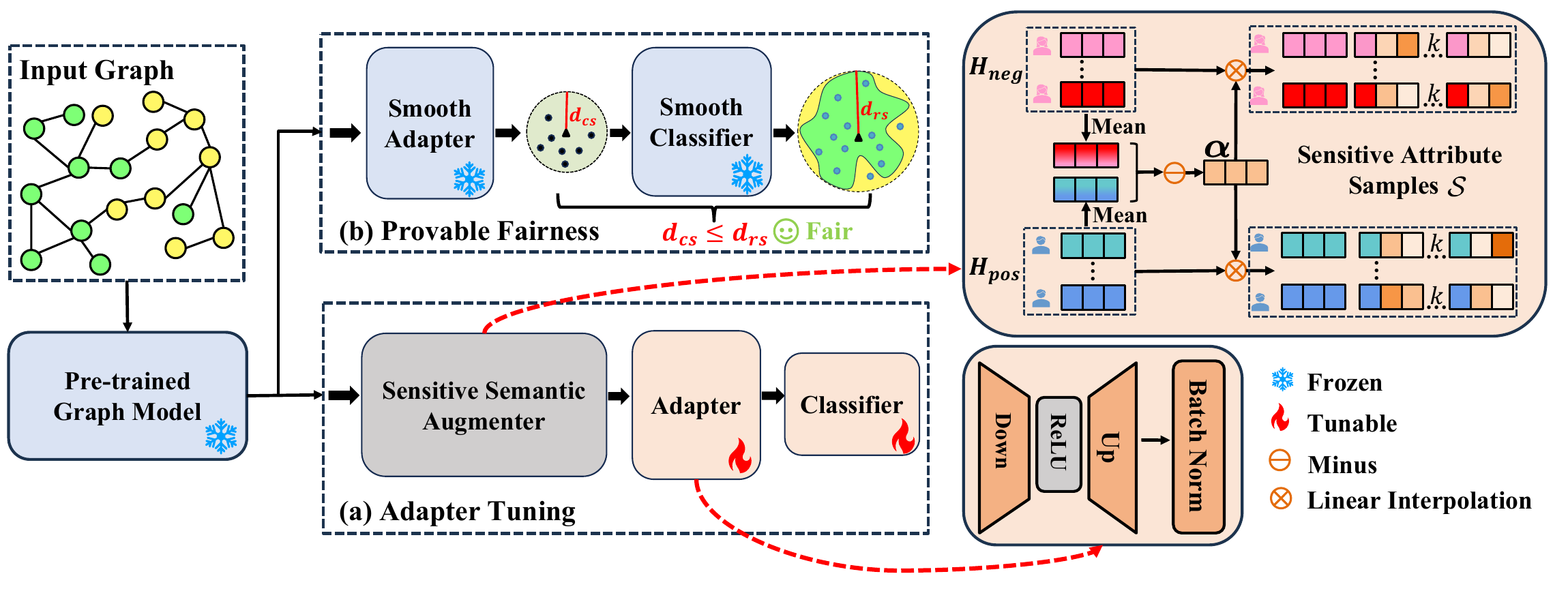}
    \vspace{-20pt}
    \caption{Overview of GraphPAR. In the adapter tuning phase, we first utilize the PGMs to obtain node representations $\mathbf{H}$. Then, we design a sensitive semantic augmenter to extend the node representations with different sensitive attribute semantics, i.e., sensitive attribute samples $\mathcal{S}$. Finally, the extended node representations are used to train an adapter, transforming the node representations to be independent of sensitive attribute semantics. In the provable fairness phase, based on the smoothed versions of the well-trained adapter and classifier, we use the smooth adapter to get its output bound guarantee $d_{cs}$ and use the smooth classifier to get its local robustness guarantee $d_{rs}$. Sequentially, we quantify whether the fairness of each node is provable by comparing $d_{cs}$ with $d_{rs}$.}
    \label{fig: framework}
    \vspace{-10pt}
\end{figure*}

In this section, we proposed a novel adapter-tuning framework, GraphPAR, which flexibly and efficiently improves the fairness of PGMs. First, we define the fairness of PGMs on GraphPAR, requiring that the predictions are not affected by the sensitive attribute semantics variations of the node. Next, to achieve the fairness objective, as depicted in Figure~\ref{fig: framework} (a) Adapter Tuning, GraphPAR consists of two key components: (1) A sensitive semantic augmenter, which extends the node representations with different sensitive attribute semantics for each node, to help further train an adapter. (2) An adapter, which transforms the node representations to be independent of sensitive attribute semantics by adversarial debiasing methods, preventing the propagation of sensitive attribute semantics from PGMs to task predictions.

\subsection{Fairness Definition of PGMs}
Given a graph $\mathcal G=(\mathcal{V},\mathcal{E},\mathbf{X})$ and the PGM $f$ pre-trained on this graph. GraphPAR freezes the parameters of $f$ to obtain node representations $\mathbf H=f(\mathcal{V},\mathcal{E},\mathbf{X})$, where $\mathbf H$ inevitably contains sensitive attribute semantics on graph $\mathcal G$ and therefore we hope utilizes an adapter $g$ to remove them. Combined with Definition~\ref{def: graph_count_fair}, the fairness of PGMs on GraphPAR is defined as follows:
\begin{definition}[Fairness of PGMs]\label{def: pgm_fair}
    In GraphPAR, a PGM $f(\cdot)$, an adapter $g(\cdot)$ and a classifier $d(\cdot)$ satisfy fairness during adaptation to downstream tasks if for any node $v_i$:
    \begin{equation}
    P((\hat{y}_i)_{\mathbf{S_h}\leftarrow{\mathbf{s}}} | \mathbf{H})=P((\hat{y}_i)_{\mathbf{S_h}\leftarrow{\mathbf{s}}^{\prime}} | \mathbf{H}), \quad \forall \mathbf{s}, \mathbf{s}^{\prime} \ \text{ s.t. } \ || \mathbf{s}-{\mathbf{s}}^{\prime} ||_2 \neq 0,
    \end{equation}
    where $\hat{y}_i= d \circ g(\mathbf{h}_i)$ denotes the predicted label for node $v_i$. $\mathbf{s}$ and $\mathbf{s}^{\prime}$ are two vectors with the same dimension as the node representation $\mathbf{h}_i$, i.e., $\mathbf{s}, \mathbf{s}^{\prime} \in \mathbb{R}^p$, representing different sensitive attribute semantics.
\end{definition}
 The Definition \ref{def: pgm_fair} means that the model prediction results remain consistent as the sensitive attribute semantics change. For example, when a person's gender semantics gradually transits from male to female, fairness is satisfied if the model predictions are always consistent, otherwise not satisfied. Thus, the objective of GraphPAR is to remove the impact of sensitive attribute semantics from $\mathbf H$ on model prediction by training an adapter $g$, improving the fairness of PGMs in downstream tasks.

\subsection{Sensitive Semantic Augmenter}
The sensitive semantic augmenter extends the node representation with different sensitive attribute semantics for each node, to help further train the adapter. Initially, according to known sensitive attribute information on the graph and representations of the nodes, we calculate a vector $\boldsymbol{\alpha}$ that represents the direction of the sensitive attribute semantics. Subsequently, we extend the node representation $\mathbf{h}_{i}$ for each node via linearly interpolating in the direction of $\boldsymbol{\alpha}$, obtaining a sensitive attribute semantics augmentation set $\mathcal{S}_i$.

\textbf{Computing the sensitive attribute semantics vector $\boldsymbol{\alpha}$.} Leveraging the capabilities of PGMs in capturing both graph structure and node attributes, we expect to derive a vector $\boldsymbol{\alpha}$ that effectively represents the sensitive attribute semantics. First, we utilize the given PGM $f$ to obtain node representations $\mathbf{H}$. Then, based on known node sensitive attribute $s$ on the graph, we partition the node representations into positive and negative sets, i.e., $\mathbf{H}_{pos}$ and $\mathbf{H}_{neg}$. Subsequently, we calculate the average representation $\mathbf{h}_{pos}$ for nodes with the sensitive attribute and $\mathbf{h}_{neg}$ for nodes without it, both obtained from $\mathbf{H}_{pos}$ and $\mathbf{H}_{neg}$, respectively. Lastly, the difference between $\mathbf{h}_{pos}$ and $\mathbf{h}_{neg}$ represents the sensitive attribute semantics vector:
\begin{align}\label{eq: vector}
    \boldsymbol{\alpha} &= \mathbf{h}_{pos} - \mathbf{h}_{neg},\\ 
    \mathbf{h}_{pos} &= \frac{1}{n_{pos}}\sum_{i=1}^{n_{pos}}\mathbf{H}_{pos,i}\ ,
    \mathbf{h}_{neg} = \frac{1}{n_{neg}}\sum_{i=1}^{n_{neg}}\mathbf{H}_{neg,i}\ ,
\end{align}
where $n_{pos}$ and $n_{neg}$ denote the number of positive and negative samples. Intuitively, the key thought in calculating $\boldsymbol{\alpha}$ is to represent the semantic relationships between groups by using the average difference in embeddings across different groups. Averaging is done to obtain a semantic embedding that represents the characteristics of a group and eliminates individual characteristic differences. For example, Mean (salesman, waiter, king) yields an embedding representing the male group, and Mean (saleswoman, waitress, queen) can produce an embedding for the female group. Subtraction is used to capture the semantic relationship between groups. For instance, man $-$ woman represents the translation vector from female to male. If a user $u$ is female, $u$ $+$ (man $-$ woman) can yield a male representation of that user. In summary, with the vector $\boldsymbol{\alpha}$, we expect to move in the direction of $\boldsymbol{\alpha}$ to increase the presence of the sensitive attribute, while moving in the opposite direction diminishes its presence. In the experiment section, we conduct detailed experiments to illustrate the effectiveness of $\boldsymbol{\alpha}$.

\textbf{Augmenting sensitive attribute semantics.} After calculating the sensitive attribute semantics vector $\boldsymbol{\alpha}$, we employ it to augment a set of sensitive attribute $\mathcal S_{i}$ for each node representation $\mathbf{h}_{i}$. This augmentation is achieved through a linear interpolation method and can be expressed as:
\begin{equation}
    \mathcal S_{i}:=\{{\mathbf h_i}+t \cdot \boldsymbol{\alpha} \mid |t| \leq \epsilon \} \subseteq \mathbb{R}^p,    
\end{equation}
where $\epsilon$ represents the augmentation range applied to the direction of the sensitive attribute semantics. The above augmentation method offers two key advantages: (1) It efficiently extends node representations with different semantics of sensitive attributes as line segments. These line segments correspond to multiple points in the original input space, bypassing complex augmentation designs in the original graph~\cite{agarwal2021towards, zhang2021multi}. (2) Although this work primarily focuses on single sensitive attribute scenarios, this method can be simply extended to situations involving multiple sensitive attributes. In such cases, interpolation can be performed along multiple sensitive attribute semantics vectors.

\subsection{Training Adapter for PGMs Fairness}
Given any PGM $f$ and the sensitive attribute augmentation set $\mathcal{S}$, we now outline how to improve the fairness of PGMs by training a parameter-efficient adapter $g$ while ensuring the prediction performance of downstream tasks. In GraphPAR, we employ two adversarial debiasing methods for training the adapter: random augmentation and min-max adversarial training.

\textbf{Random augmentation adversarial training (RandAT).} 
During the adapter $g$ training process, we choose $k$ samples from the augmented sensitive attribute set ${\mathcal S_{i}}$ to obtain adversarial training set $\hat{\mathcal S_{i}}$, i.e.,
\begin{equation}\label{eq: set}
\hat{\mathcal S_{i}} = \{{\mathbf h_i}+t_j \cdot \boldsymbol{\alpha}\}_{j=1}^k,\ t_j \sim \text{Uniform}(-\epsilon, \epsilon),
\end{equation} 
where $\epsilon$ represents the augmentation range. These selected samples are then incorporated into the training of the adapter. The optimization loss can be formulated as follows:
\begin{equation}\label{eq: dataaug}
    \mathcal{L}_{\mathrm{RandAT}}=
    \mathbb E_{i \in \mathcal V_L}\left[
        \mathbb E_{\mathbf{h}_{i}^{\prime} \in  \hat{\mathcal S_{i}}}\left[
            \ell (d \circ g(\mathbf{h}_{i}^{\prime}),y_i)
        \right]
    \right],
\end{equation}
where $\mathcal V_L$ is the set of labeled nodes, $d$ is a downstream classifier, and $\ell (\cdot) $ is cross-entropy loss which measures the prediction error.

In RandAT, by introducing diverse sensitive attribute semantic samples in the training process, the adapter $g$ and the classifier $d$ become more robust to variations in sensitive information, mitigating potential discriminatory predictions. At the same time, these augmented samples share the same task-related semantics as the original sample, which further helps the adapter and classifier capture task-related semantics.

\textbf{Min-max adversarial training (MinMax).}
Unlike RandAT, the key thought behind MinMax is to find and optimize the worst-case in each round. Our objective is to minimize the discrepancy between the representation $\mathbf{h}_{i}$ and its corresponding augmented sensitive attribute semantics set ${\mathcal S_{i}}$. This is achieved by ensuring that the representation $\mathbf{h}_i$ closely aligns with the representations within ${\mathcal S_{i}}$. To quantify this alignment, we seek to minimize the distance between $\mathbf{h}_{i}$ and ${\mathcal S_{i}}$. The optimization objective of MinMax is minimizing the following loss function:

\begin{equation}\label{eq: random-adv}
\mathcal{L}_{MinMax}\left(\mathbf{h}_{i}\right)=\max_{\mathbf{h}_{i}^{\prime}\in \mathcal S_{i}}\left\|g\left(\mathbf{h}_{i}\right)-g\left(\mathbf{h}_{i}^{\prime}\right)\right\|_{2},
\end{equation}
where $\mathbf{h}_{i}^{\prime}$ have different sensitive attribute semantics with $\mathbf{h}_{i}$. Minimizing $\mathcal L_{adv}(\mathbf{h}_i)$ is a min-max optimization problem, and adversarial training is effective in this scenario. Since the input domain of the inner maximization problem is a simple line segment about $\boldsymbol{\alpha}$, we can perform adversarial training ~\cite{engstrom2019exploring} by uniformly sampling $k$ points from $\mathcal S_{i}$ to construct $\hat{\mathcal S_{i}}$ and approximate it as follow:

\begin{equation}
\mathcal{L}_{MinMax}\left(\mathbf{h}_{i}\right)\approx\max_{\mathbf{h}_{i}^\prime \in \hat{\mathcal S_{i}}}\left\Vert g\left(\mathbf{h}_{i}\right)-g\left({\mathbf h_{i}^{\prime}}\right)\right\Vert_{2}.
\end{equation}

To further ensure that the adapter $g$ does not filter out useful task information, we introduce cross-entropy classification loss to ensure the performance of the downstream task:

\begin{equation}
    \mathcal{L}_{cls}\left(\mathbf{h}_{i},y_i\right)=\ell({d \circ f(\mathbf{h}_{i})}, y_i).
\end{equation}

The final optimization objective of MinMax is as follows:
\begin{equation} \label{eq: minmax}
    \mathcal{L}=\lambda \mathcal{L}_{MinMax} +  \mathcal{L}_{cls},
\end{equation}
where $\lambda$ is a scale factor for balancing accuracy and fairness.

\section{Provable Fair Adaptation of PGMs}
In this section, based on GraphPAR, we primarily discuss how to provide provable fairness for each node, i.e., the prediction results are consistent within a certain range of sensitive attribute semantics. We divide this process into two key components as depicted in Figure \ref{fig: framework} (b) Provable Fairness: (1) Smooth adapter. We construct a smoothed version for the adapter using center smooth, which provides a bound for the output variation of node representation $\mathbf{h}$ within the range of sensitive attribute semantics change. This guarantees that the range of output results is contained within a minimal enclosing ball centered at $\mathbf z$ with a radius of $d_{cs}$. (2) Smooth classifier. We construct a smoothed version for the classifier using random smooth, which provides local robustness against the center $\mathbf{z}$. By determining whether all points within the minimum enclosing ball are classified into the same class, i.e., $d_{cs} < d_{rs}$, we quantify if the fairness of each node is provable. Note that based on the well-trained adapter and classifier, the smoothed models are constructed by different definitions of the smoothing function. Thus, the process of construction does not require training in any parameters. We leave out the subscript $(\cdot)_{i}$ for notation simplicity.

\subsection{Provable Adaptation}

To guarantee the range of change in the representation after applying the adapter $g$, we employ center smoothing ~\cite{kumar2021center} to obtain a smoothed version of the adapter, denoted as $\widehat g$. It provides a guarantee for the output bound of the adapter with a representation $\mathbf h$ as the input, described in Theorem~\ref{th: center}: 
\begin{theorem}[Center Smoothing ~\cite{kumar2021center}] \label{th: center}
Let $\widehat g$ denote an approximation of the smoothed version of the adapter $g$, which maps a representation $\mathbf h$ to the center point $\widehat g(\mathbf h)$ of a minimum enclosing ball containing at least half of the points $\mathbf{z}\sim g(\boldsymbol{\mathbf h}+\mathcal{N}(0,\sigma_{cs}^2I))$. The formal definition of $\widehat g$ as follows:
\begin{equation} \label{eq:def-center}
    \widehat{g}{(\mathbf{h})}=\underset{\mathbf z}{\operatorname*{argmin}} \ r \ \text{s.t.} \ \mathbb{P}[g(\boldsymbol{\mathbf h}+\mathcal{N}(0,\sigma_{cs}^2I))\in\mathcal{B}(\mathbf{z},r)]\geq\frac{1}{2}, 
\end{equation}
where $\mathbf{z}$ and $r$ are the center and radius of the minimum enclosing ball, respectively. Then, for an $l_2$-perturbation size $\epsilon_1 >0$ on $\mathbf h$, we can produce a guarantee $d_{cs}$ of the output change with confidence $1 - {\alpha}_{cs}$:
\begin{equation}
    \forall \ \boldsymbol{\mathbf h}^{\prime} \ s.t.\ \|\boldsymbol{\mathbf h}-\boldsymbol{\mathbf h}^{\prime}\|_2\leq \epsilon_1 \ ,\|\widehat{g}(\boldsymbol{\mathbf h})-\widehat{g}(\boldsymbol{\mathbf h}^{\prime})\|_2\leq d_{cs},
\end{equation}
since $\mathbf h^{\prime} - \mathbf h = t \cdot \boldsymbol{\alpha} $, we have:
\begin{equation}
\epsilon_1=\max_{t}\|t\boldsymbol{\alpha}\|=\epsilon\|\boldsymbol{\alpha} \|_2,
\end{equation}
where $\epsilon$ represents the augmentation range applied to the direction of the sensitive attribute semantics. 
\end{theorem}
 Theorem~\ref{th: center} implies that given a node representation $\mathbf h$ and its set of sensitive attribute samples $\mathcal S$ with range $\epsilon$, a guarantee $d_{cs}$ can be computed with high probability. This $d_{cs}$ represents the range of the adapter output changes, serving as a meaningful certificate. It guarantees that when the sensitive attribute semantics of input $\mathbf h$ is perturbed within a range defined by $\epsilon$, the range of output remains within a minimal enclosing ball.

\subsection{Provable Classification}

Next, it is necessary to demonstrate that predictions for all points within this minimum enclosing ball are classified consistently. This consistency guarantees the effectiveness of debiasing results.

\begin{theorem}\label{th: random}
    (Random Smoothing ~\cite{cohen2019certified}) Let $d$ be a classifier and let $\mathbf{\varepsilon} \sim \mathcal{N}(0,\sigma_{rs}^2I)$. The smoothing version of the classifier $\widehat d$ is defined as follows:
    \begin{equation} \label{eq:def-random}
        \widehat{d}\left(\mathbf{z}\right)=\arg\max_{y\in\mathcal{Y}}\mathbb{P}_{{\mathbf{\varepsilon}}}(d(\mathbf{z}+\mathbf{\varepsilon})=y).
    \end{equation}
Suppose $y_A \in \mathcal{Y}$ and $\underline {p_A}, \overline{p_B} \in [0, 1]$  satisfy:
\begin{equation}
    \mathbb{P}_{\mathbf{\varepsilon}}(d(\mathbf{z}+\mathbf{\varepsilon})=y_A)\geq\underline{p_A}\geq\overline{p_B}\geq\max_{y_B\neq y_A}\mathbb{P}_{\mathbf{\varepsilon}}(d(\mathbf{z}+\mathbf{\varepsilon})=y_B).
\end{equation}
Then, we have $\widehat{d}(\mathbf{z}+\mathbf{\delta})=y_A$ for all $\mathbf{\delta}$ satisfying $\|\mathbf{\delta}\|_2<d_{rs}$, where $d_{rs}$ can be obtain as follow:
\begin{equation}
    d_{rs}:=\frac{\sigma_{rs}}2(\Phi^{-1}(\underline{p_A})-(\Phi^{-1}(\overline{p_B})),
\end{equation}
where $\mathcal Y$ denotes the set of class labels, $\Phi$ is the cumulative distribution function (CDF) of the standard normal distribution $\mathcal N (0, 1)$, and $\Phi^{-1}$ is its inverse.
\end{theorem}

Theorem~\ref{th: random} derives a local robustness radius $d_{rs}$ for the input $\mathbf{z}$ by employing the smoothed version $\widehat d$ of the classifier $d$. This robustness guarantees that within the verified region of input, which is bounded by $d_{rs}$, the classification output of $\widehat d$ remains consistent, providing a guarantee of stability and consistency in the prediction results. Theorem~\ref{th: random} is especially important for providing provable fairness, because if $d_{cs} < d_{rs}$, then it guarantees consistency in the predictions to different sensitive attribute semantic samples.

\subsection{GraphPAR Provides Provable Fairness}
To establish a theoretical guarantee for the debiasing effect of adapter $g$, combining Theorem~\ref{th: center} and Theorem~\ref{th: random}, we define the provable fairness of PGMs as follows:
\begin{definition}[Provable Fairness of PGMs]\label{def: provable-fair}
Given a node representation $\mathbf h$, the debiasing process $M$ satisfies:
\begin{equation}
    M(\mathbf h) = M(\mathbf h^\prime), \forall \ \mathbf h^\prime \in \mathcal S\ ,
\end{equation}
where $\mathcal S$ is the set of sensitive attribute augmentations for $\mathbf h$.
\end{definition}
With the aforementioned two smoothing techniques, the provable fairness of PGMs is naturally achieved with the following theorem:
\begin{theorem}\label{th: provable-fairness}
Assuming we have a PGM $f$, a center smoothing adapter $\widehat g$, and a random smoothing classifier $\widehat d$. For the $i$-th node, if $\widehat g$ obtains a output guarantee $d_{cs}$ with confidence $1-{\alpha}_{cs}$ and $\widehat d$ obtains a local robustness guarantee $d_{rs}$ with confidence $1-{\alpha}_{rs}$, and satisfy $d_{cs} < d_{rs}$, then the fairness of the debiasing $M=\widehat{d}\circ\widehat{g}\circ f(\mathcal{V},\mathcal{E},\mathbf{X})_i$ is provable with a confidence $1-{\alpha}_{cs}-{\alpha}_{rs}$, the definition of the formalization is as follows:
\begin{equation}\label{eq: def}
    \forall \ \mathbf{h}^{\prime}\in \mathcal{S}:\widehat{d}\circ\widehat{g}\left(\mathbf{h}\right)=\widehat{d}\circ\widehat{g}\left(\mathbf{h}^{\prime}\right),
\end{equation}
where $\mathbf{h} \in \mathbf{H}, \mathbf H=f(\mathcal{V},\mathcal{E},\mathbf{X})$.
\end{theorem}
The detailed algorithms process of Theorem~\ref{th: provable-fairness} is referred to Appendix~\ref{algo: graphpar}, and the proof of Theorem~\ref{th: provable-fairness} is referred to Appendix~\ref{proof: provable-fairness}.

\section{Experiments}
In this section, we extensively evaluate GraphPAR to answer the following research questions (RQs). 
\textbf{RQ1}: How effective is GraphPAR compared to existing graph fairness methods? 
\textbf{RQ2}: Compared to methods without debiasing adaptation, does GraphPAR show improvement in the number of nodes with provable fairness?
\textbf{RQ3}: How effective is the vector $\boldsymbol{\alpha}$ in representing the sensitive attribute semantics direction?
\textbf{RQ4}: How do different hyperparameters of GraphPAR impact the classification performance and fairness? 
\textbf{RQ5}: How parameter-efficient is GraphPAR?

\textbf{Experimental setup.} We test GraphPAR on the node classification task. These are common graph datasets with sensitive attributes collected from various domains. We choose four public datasets \textit{Income}~\cite{asuncion2007uci}, \textit{Credit}~\cite{agarwal2021towards}, \textit{Pokec\_z} and \textit{Pokec\_n}~\cite{dai2021say}. Datasets statistics refer to Table~\ref{tab:datasets} and more details refer to Appendix~\ref{exp: detail-dataset}. Implementation details of GrahPAR refer to Appendix~\ref{exp: exp-set}. We report the experiment results over five runs with different random seeds. 

\begin{table}[tbp]
\Huge
\centering
\caption{Datasets Statistics.}
\vspace{-5pt}
\label{tab:datasets}
\begin{adjustbox}{width=0.46\textwidth}
\begin{tabular}{@{}lllll@{}}
\toprule
\textbf{Dataset}             & \textbf{Credit}         & \textbf{Pokec\_n}      & \textbf{Pokec\_z}      & \textbf{Income}       \\ \midrule
\textbf{\#Nodes}             & 30,000         & 66,569        & 67,797        & 14,821       \\
\textbf{\#Features}          & 13             & 266           & 277           & 14           \\
\textbf{\#Edges}             & 1,436,858      & 729,129       & 882,765       & 100,483      \\
\textbf{Node label}          & Future default & Working field & Working field & Income level \\
\textbf{Sensitive attribute} & Age            & Region        & Region        & Race         \\
\textbf{Avg. degree}         & 95.79          & 16.53         & 19.23         & 13.6         \\ \bottomrule
\end{tabular}
\end{adjustbox}
\vspace{-15pt}
\end{table}

\textbf{Baselines. }We compare GraphPAR to four baselines: vanilla GCN~\cite{kipf2016semi}, graph fairness methods FairGNN~\cite{dai2021say}, NIFTY~\cite{agarwal2021towards}, and EDITS~\cite{dong2022edits}. We choose contrastive pre-training DGI~\cite{DGI2018} and GCA~\cite{zhu2021graph}, as well as predictive pre-training EdgePred~\cite{hamilton2017inductive} as the backbone of GraphPAR. More details of baselines refer to Appendix~\ref{exp: detail-baseline}.

\subsection{Prediction Performance and Fairness (RQ1)} 
\renewcommand{\arraystretch}{1.1}
\begin{table*}[htbp]
 \caption{Performance and fairness ($\%\pm\sigma$) on node classification. The best results are in bold and runner-up results are underlined.} 
 \vspace{-5pt}
 \label{tab:node-classification}
 \begin{adjustbox}{width=\textwidth}
\begin{tabular}{@{}cccccccccccccc@{}}
\toprule
\multicolumn{2}{c}{\multirow{2}{*}{\Large{Method}}} &
  \multicolumn{4}{c}{\large{Credit}} &
  \multicolumn{4}{c}{\large{Pokec\_z}} &
  \multicolumn{4}{c}{\large{Pokec\_n}} \\ \cmidrule(l){3-14} 
\multicolumn{2}{c}{} &
  ACC (↑) &
  F1 (↑) &
  DP (↓) &
  EO (↓) &
  ACC (↑) &
  F1 (↑) &
  DP (↓) &
  EO (↓) &
  ACC (↑) &
  F1 (↑) &
  DP (↓) &
  EO (↓) \\ \midrule
\multicolumn{2}{c}{GCN} &
  \large{69.73±0.04} &
  \large{79.14±0.02} &
  \large{13.28±0.15} &
  \large{12.66±0.24} &
  \large{67.54±0.48} &
  \large{68.93±0.39} &
  \large{5.51±0.67} &
  \large{4.57±0.29} &
  \textbf{\large{70.11±0.34}} &
  {\ul \large{67.37±0.38}} &
  \large{3.19±0.86} &
  \large{2.93±0.95} \\
\multicolumn{2}{c}{FairGNN} &
  \large{72.50±4.09} &
  \large{81.80±3.86} &
  \large{9.20±3.35} &
  \large{7.64±3.58} &
  \large{67.47±1.12} &
  \large{69.35±3.14} &
  \large{1.91±1.01} &
  \large{1.04±1.11} &
  \large{68.42±2.04} &
  \large{64.34±2.32} &
  \large{1.41±1.30} &
  \large{1.50±1.23} \\
\multicolumn{2}{c}{NIFTY} &
  \large{70.89±0.59} &
  \large{80.23±0.54} &
  \large{9.93±0.59} &
  \large{8.79±0.71} &
  \large{65.83±3.90} &
  \large{66.99±4.26} &
  \large{5.47±2.13} &
  \large{2.64±1.02} &
  \large{68.97±1.21} &
  \large{66.77±1.27} &
  \large{1.68±0.90} &
  \large{1.38±0.91} \\
\multicolumn{2}{c}{EDITS} &
  \large{66.80±1.03} &
  \large{76.64±1.13} &
  \large{10.21±1.14} &
  \large{8.78±1.15} &
  \large{OOM} &
  \large{OOM} &
  \large{OOM} &
  \large{OOM} &
  \large{OOM} &
  \large{OOM} &
  \large{OOM} &
  \large{OOM} \\ \midrule
\multirow{3}{*}{\large{DGI}} &
  Naive &
  {\ul \large{75.72±2.18}} &
  {\ul \large{84.73±2.00}} &
  \large{7.87±2.22} &
  \large{6.51±2.79} &
  {\ul \large{67.87±0.51}} &
  \large{70.23±0.80} &
  \large{4.69±1.95} &
  \large{3.03±1.34} &
  \large{68.58±1.22} &
  \large{65.66±1.37} &
  \large{3.58±3.09} &
  \large{4.99±3.68} \\
 &
  $\text{GraphPAR}_{RandAT}$ &
  \textbf{\large{76.88±1.33}} &
  \textbf{\large{85.85±1.36}} &
  \large{5.93±2.91} &
  \large{4.44±3.34} &
  \large{67.05±1.33} &
  \textbf{\large{70.50±0.69}} &
  \large{1.90±1.22} &
  {\ul \large{0.84±0.28}} &
  \large{68.92±1.55} &
  \large{65.61±1.33} &
  \textbf{\large{1.19±0.65}} &
  \large{2.11±1.60} \\
 &
  $\text{GraphPAR}_{MinMax}$ &
  \large{74.37±2.91} &
  \large{83.46±2.64} &
  \textbf{\large{3.81±2.37}} &
  \textbf{\large{2.60±2.48}} &
  \textbf{\large{68.32±0.55}} &
  \large{68.35±2.38} &
  {\ul \large{1.64±0.78}} &
  \textbf{\large{0.53±0.39}} &
  \large{68.43±0.55} &
  \textbf{\large{68.20±2.22}} &
  \large{1.73±0.76} &
  {\ul \large{1.11±0.88}} \\ \midrule
\multirow{3}{*}{\large{EdgePred}} &
  Naive &
  \large{69.66±1.74} &
  \large{79.30±1.63} &
  \large{7.89±2.28} &
  \large{6.67±2.42} &
  \large{67.33±0.44} &
  \large{69.17±0.52} &
  \large{6.00±3.04} &
  \large{3.95±2.52} &
  \large{68.60±0.53} &
  \large{65.56±0.79} &
  \large{2.48±0.86} &
  \large{5.29±2.71} \\
 &
  $\text{GraphPAR}_{RandAT}$ &
  \large{69.97±2.35} &
  \large{79.55±2.24} &
  \large{6.36±2.19} &
  \large{4.83±2.70} &
  \large{66.87±1.12} &
  \large{68.86±0.46} &
  \large{1.99±1.12} &
  \large{2.27±1.23} &
  \large{68.49±1.41} &
  \large{65.45±1.02} &
  \large{1.79±0.85} &
  \large{3.69±0.68} \\
 &
  $\text{GraphPAR}_{MinMax}$ &
  \large{68.53±1.23} &
  \large{78.19±1.14} &
  \large{5.10±2.31} &
  \large{4.52±2.17} &
  \large{67.51±0.55} &
  \large{69.03±0.82} &
  \textbf{\large{1.45±1.40}} &
  \large{1.15±0.85} &
  {\ul \large{69.10±0.91}} &
  \large{65.00±1.10} &
  {\ul \large{1.28±0.97}} &
  \large{3.31±2.06} \\ \midrule
\multirow{3}{*}{\large{GCA}} &
  Naive &
  \large{75.28±0.51} &
  \large{84.35±0.47} &
  \large{8.56±0.97} &
  \large{6.21±0.90} &
  \large{67.63±0.44} &
  \large{70.24±0.98} &
  \large{7.68±2.19} &
  \large{4.82±1.43} &
  \large{67.85±1.23} &
  \large{65.81±1.35} &
  \large{2.90±2.61} &
  \large{3.23±1.05} \\
 &
  $\text{GraphPAR}_{RandAT}$ &
  {\large{75.50±1.29}} &
  {\large{84.66±1.27}} &
  \large{5.51±2.44} &
  \large{3.98±1.96} &
  \large{66.73±2.22} &
  {\ul \large{70.32±0.73}} &
  \large{4.23±2.50} &
  \large{2.94±1.84} &
  \large{68.11±0.44} &
  \large{64.43±1.05} &
  \large{2.35±1.12} &
  \large{2.42±1.62} \\
 &
  $\text{GraphPAR}_{MinMax}$ &
  \large{73.74±2.01} &
  \large{82.96±1.74} &
  {\ul \large{4.90±1.90}} &
  {\ul \large{2.96±1.66}} &
  \large{66.59±1.28} &
  \large{68.74±1.17} &
  \large{2.33±2.28} &
  \large{2.42±1.72} &
  \large{68.11±0.70} &
  \large{65.49±1.57} &
  \large{1.41±0.86} &
  \textbf{\large{0.94±0.59}} \\ \bottomrule
\end{tabular}
\end{adjustbox}
\vspace{-10pt}
\end{table*}
We choose accuracy (ACC) and macro-F1 (F1) to measure how well the nodes are classified, demographic parity (DP) and equality of opportunity (EO) to measure how fair the classification is. The results are shown in Table~\ref{tab:node-classification}; additional results on Income refer to Table~\ref{tab:income-node-classification} in Appendix~\ref{exp: effect-income}. We interpret the results as follows:\\
$\bullet$ GraphPAR outperforms baseline models both in classification and fairness performance. GraphPAR is demonstrated to be superior in both classification and fairness performances, enhancing existing PGM models and outperforming other graph fairness methods. This result supports the effectiveness of GraphPAR in addressing fairness issues in the embedding space:
(1) Powerful pre-training strategies enable the embeddings to include intrinsic information for downstream tasks.
(2) Since PGMs also capture sensitive attribute information, the sensitive semantics vector can be effectively constructed.
(3) Augmenting in the embedding space is independent of task labels; thus, the sensitive semantic augmenter does not corrupt the downstream performance.\\
$\bullet$ Performance of GraphPAR varies among different PGMs. The prediction performance and fairness vary when choosing different PGMs as the backbone. Usually, we observe that contrastive pre-training methods DGI and GCA perform better than the predictive method EdgePred, implying the performance of GraphPAR is positively related to the semantic capture ability of PGMs.\\
$\bullet$ RandAT and MinMax perform well but in different ways. It is worth mentioning that RandAT often achieves the best result on classification while MinMax often performs the best on fairness. The following differences in the training schemes directly lead to the result above: 
(1) In downstream classification loss, RandAT utilizes all augmented samples, while MinMax only utilizes the original sample. As a result, RandAT often outperforms MinMax on classification metrics ACC and F1. We regard that classification benefits from data augmentation~\cite{dataaug}, as these augmented samples share the same task-related semantics as the original samples, which helps the adapter and classifier further capture task-related semantics.
(2) To debias sensitive information, MinMax minimizes the largest distance between an individual $\mathbf h_i$ and other samples $\mathbf h_{i}^{\prime}$ in the sensitive augmentation set $\mathcal{S}_i$, which can achieve a better debiasing result against the sampling strategy in RandAT that performs adversarial training on all augmented samples.

These empirical findings straightforwardly demonstrate the characteristics of RandAT in Equation~\ref{eq: dataaug} and MinMax in Equation~\ref{eq: random-adv}.

\subsection{Debiasing Guarantee (RQ2)} 
To additionally guarantee how fair the classification is, we evaluate the provable fairness of GraphPAR compared with methods without debiasing adaptation, i.e., naive PGMs. Here, with the smoothed adapter and classifier, the metrics are accuracy (ACC) and provable fairness (Prov\_Fair) in Definition~\ref{def: provable-fair}. The result is presented in Table~\ref{tab:cert-fair}; additional results on Income refer to Table~\ref{tab:income-cert-fair} in Appendix~\ref{exp: effect-income}. We have the following observations:\\
$\bullet$ Different from naive PGMs that show little or nearly zero provable fairness, RandAT achieves much better provable fairness, and MinMax has its fairness guaranteed very well. According to Theorem~\ref{th: provable-fairness} where the provable fairness of PGMs satisfies $d_{cs}<d_{rs}$, since $d_{rs}$ is the same, but $d_{cs}$ is different among training schemes: naive PGMs do not optimize $d_{cs}$, thus the fairness is nearly not guaranteed; RandAT conduct adversarial training by using many samples with different sensitive attribute semantics, which has a positive effect on minimizing $d_{cs}$ but not in an explicit way; MinMax achieves the best provable fairness by directly finding and optimizing $d_{cs}$ with min-max training.\\
$\bullet$ The classification performances of RandAT and MinMax are competitive to naive PGMs. On the one hand, RandAT does not lose its classification performance because its augmentation is performed in sensitive semantics and does not introduce noise to task-related information; on the other hand, MinMax trains the downstream classifier with original data, implying that the adapter almost has no adverse effect on the classification while guaranteeing fairness.

In conclusion, the empirical results above support that when trained with RandAT and MinMax, GraphPAR guarantees fairness without compromising its classification performance.
\renewcommand{\arraystretch}{1.5}
\begin{table}[htbp]
\caption{Provable fairness under different training schemes.}
\vspace{-7pt}
\label{tab:cert-fair}
\begin{adjustbox}{width=0.45\textwidth}
\begin{tabular}{@{}cccccccc@{}}
\toprule
\multirow{2}{*}{\huge{Dataset}} & \multirow{2}{*}{\huge{PGM}} & \multicolumn{2}{c}{\huge{Naive}} & \multicolumn{2}{c}{\huge{$\text{GraphPAR}_{RandAT}$}} & \multicolumn{2}{c}{\huge{$\text{GraphPAR}_{MinMax}$}} \\ \cmidrule(l){3-8} 
&          & \Large{ACC (↑)}        & \Large{Prov\_Fair (↑)} & \Large{ACC (↑)}        & \Large{Prov\_Fair (↑)} & \Large{ACC (↑)} & \Large{Prov\_Fair (↑)} \\ \midrule
\multicolumn{1}{c|}{\multirow{3}{*}{\huge{Credit}}}   & \Large{DGI}      & \huge{72.80}          & \huge{27.63}          & \textbf{\huge{75.39}} & \huge{37.05}          & \huge{72.71}   & \textbf{\huge{89.59}} \\
\multicolumn{1}{c|}{}                          & \Large{EdgePred} & \huge{66.87}          & \huge{5.41}           & \textbf{\huge{67.02}} & \huge{44.20}          & \huge{66.41}   & \textbf{\huge{96.28}} \\
\multicolumn{1}{c|}{}                          & \Large{GCA}      & \huge{72.86}          & \huge{0.28}           & \textbf{\huge{73.25}} & \huge{20.26}          & \huge{70.10}   & \textbf{\huge{92.92}} \\ \midrule
\multicolumn{1}{c|}{\multirow{3}{*}{\huge{Pokec\_z}}} & \Large{DGI}      & \textbf{\huge{67.30}} & \huge{1.47}           & \huge{67.21}          & \huge{10.99}          & \huge{67.28}   & \textbf{\huge{94.51}} \\
\multicolumn{1}{c|}{}                          & \Large{EdgePred} & \huge{66.02}          & \huge{0}              & \huge{66.27} & \huge{37.51}          & \textbf{\huge{66.80}}   & \textbf{\huge{90.97}} \\
\multicolumn{1}{c|}{}                          & \Large{GCA}      & \textbf{\huge{66.92}} & \huge{13.9}           & \huge{66.67}          & \huge{16.14}          & \huge{65.22}   & \textbf{\huge{95.75}} \\ \midrule
\multicolumn{1}{c|}{\multirow{3}{*}{\huge{Pokec\_n}}} & \Large{DGI}      & \textbf{\huge{68.45}} & \huge{0.70}           & \huge{67.52}          & \huge{0.52}           & \huge{68.38}   & \textbf{\huge{77.97}} \\
\multicolumn{1}{c|}{}                          & \Large{EdgePred} & \huge{67.58}          & \huge{0}              & \textbf{\huge{68.15}} & \huge{21.17}          & \textbf{\huge{68.15}}   & \textbf{\huge{88.76}} \\
\multicolumn{1}{c|}{}                          & \Large{GCA}      & \huge{67.49}          & \huge{17.80}           & \textbf{\huge{67.52}} & \huge{10.03}          & \huge{67.30}   & \textbf{\huge{91.16}} \\ \bottomrule
\end{tabular}
\end{adjustbox}
\vspace{-15pt}
\end{table}

\vspace{-5pt}
\begin{figure*}[htbp]
    \centering
    \begin{subfigure}[b]{0.23\linewidth}
        \includegraphics[width=\linewidth]{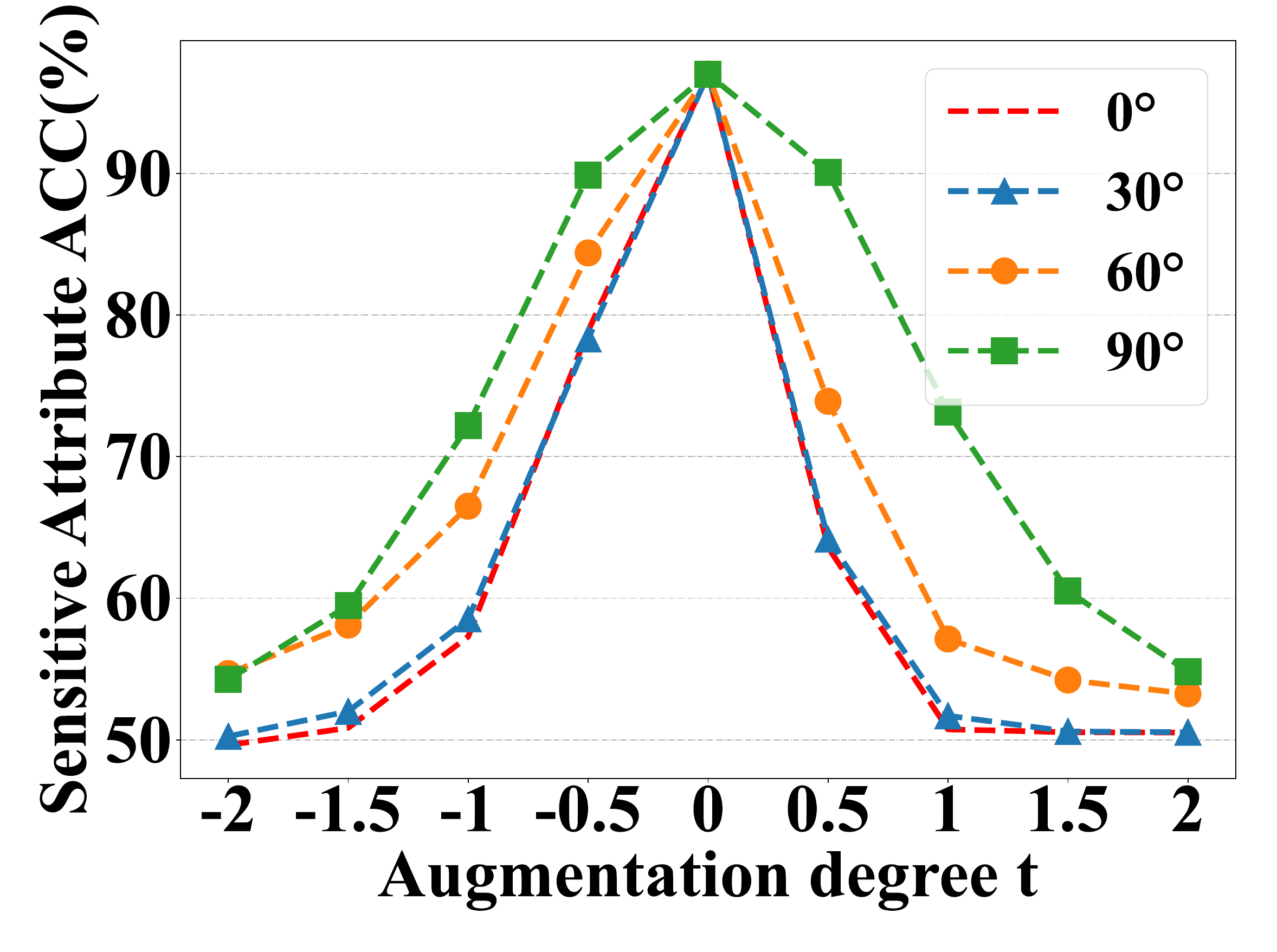}
        \vspace{-15pt}
        \caption{Pokec\_n.}
    \end{subfigure}
    \begin{subfigure}[b]{0.23\linewidth}
        \includegraphics[width=\linewidth]{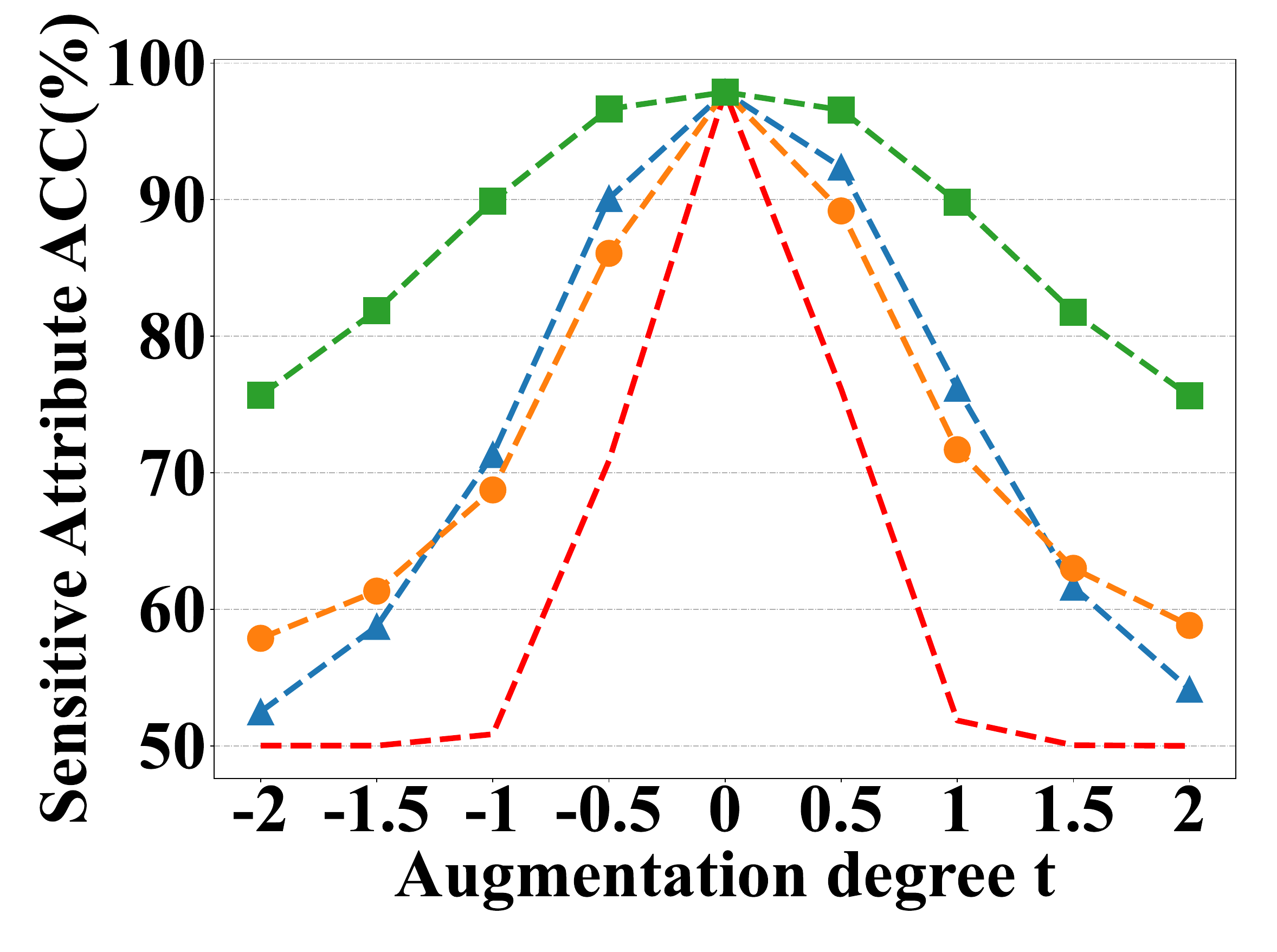}
        \vspace{-15pt}
        \caption{Pokec\_z.}
    \end{subfigure}
    \begin{subfigure}[b]{0.23\linewidth}
        \includegraphics[width=\linewidth]{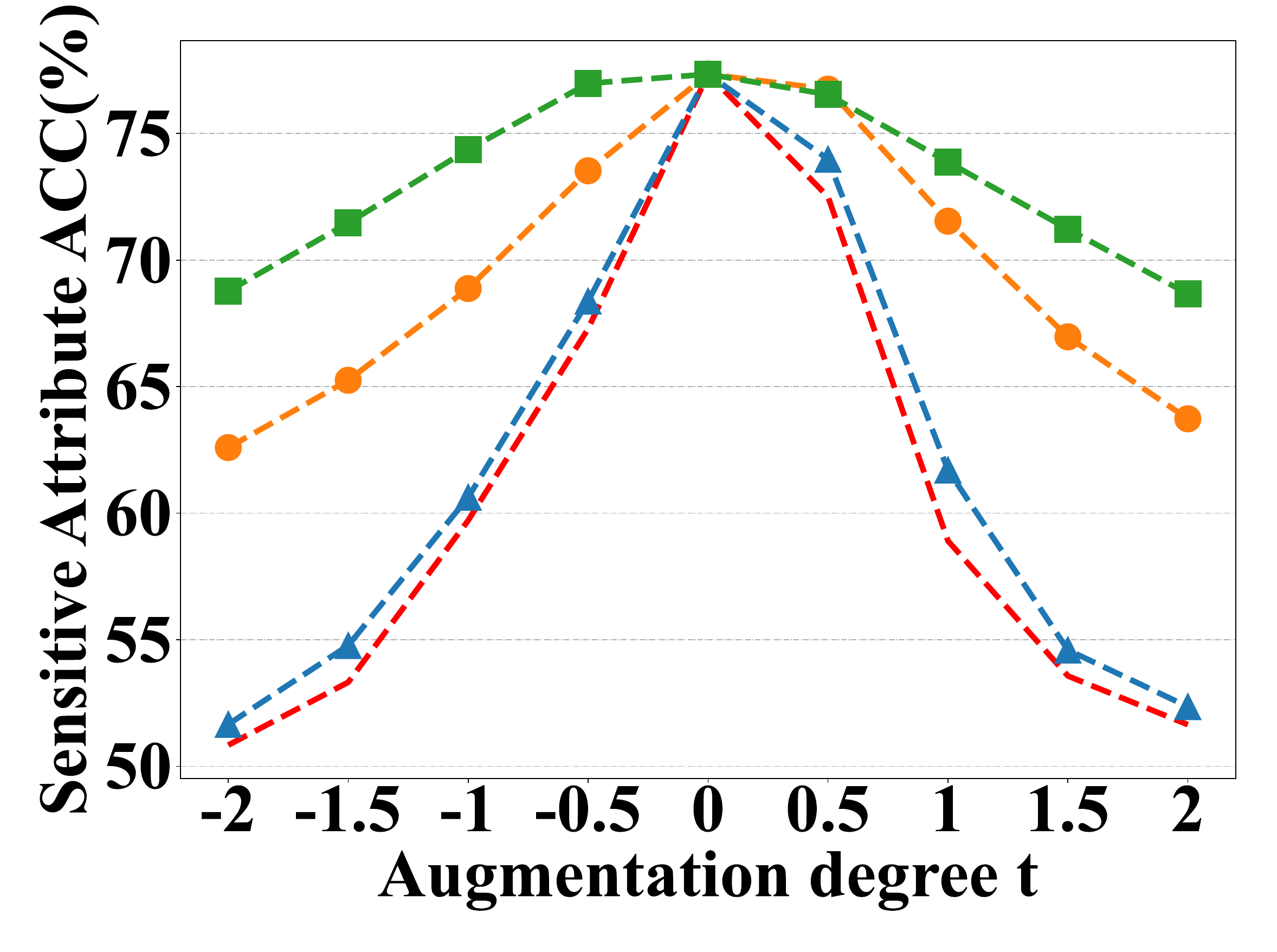}
        \vspace{-15pt}
        \caption{Credit.}
    \end{subfigure}
    \begin{subfigure}[b]{0.23\linewidth}
        \includegraphics[width=\linewidth]{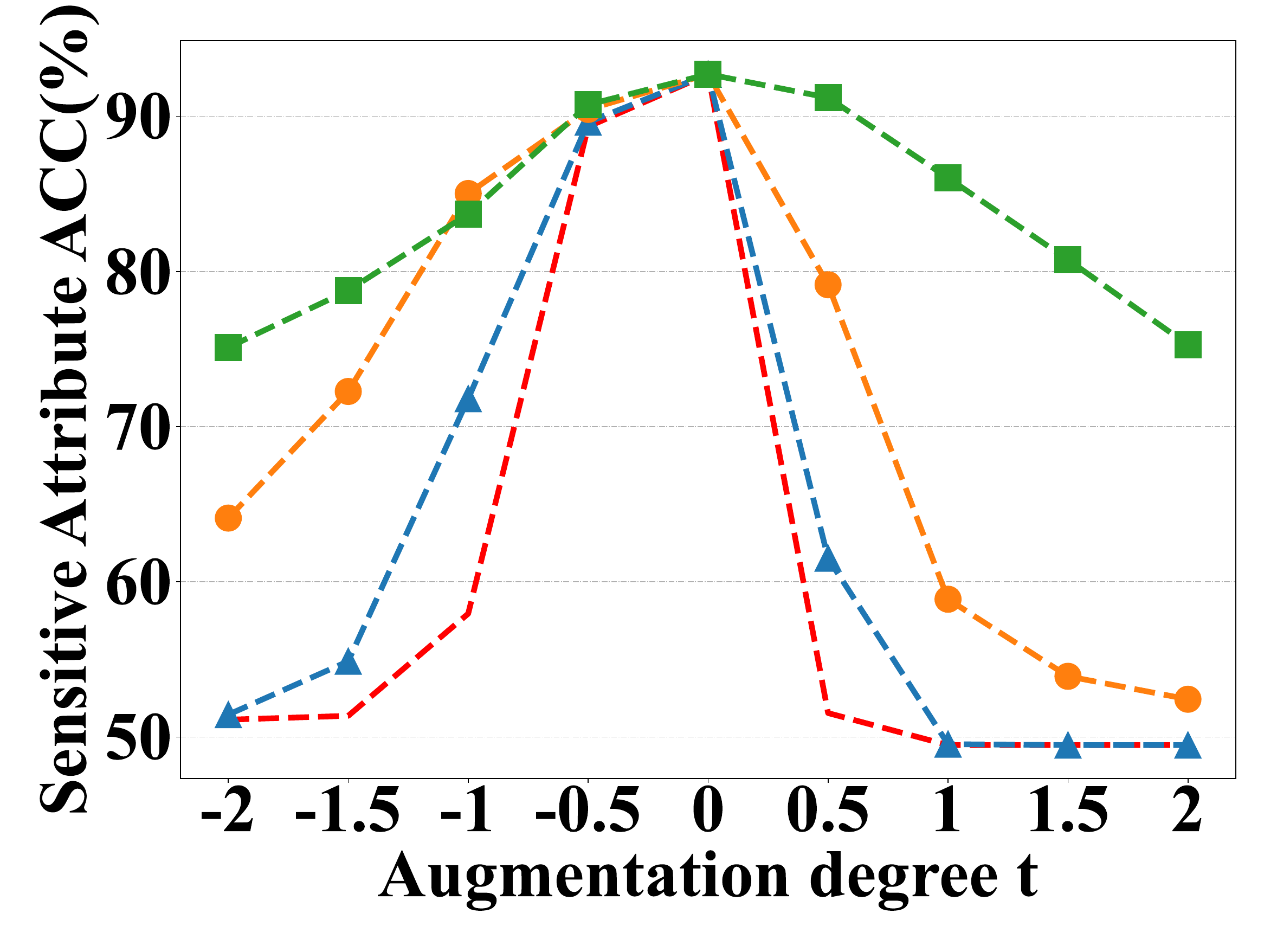}
        \vspace{-15pt}
        \caption{Income.}
    \end{subfigure}
    \vspace{-10pt}
    \caption{Sensitive attribute prediction accuracy under different augmentation degree $t$.}
    \label{fig:sens_acc}
    \vspace{-10pt}
\end{figure*}
\begin{figure*}[htbp]
    \centering
    \begin{subfigure}[b]{0.48\linewidth}
        \includegraphics[width=\linewidth]{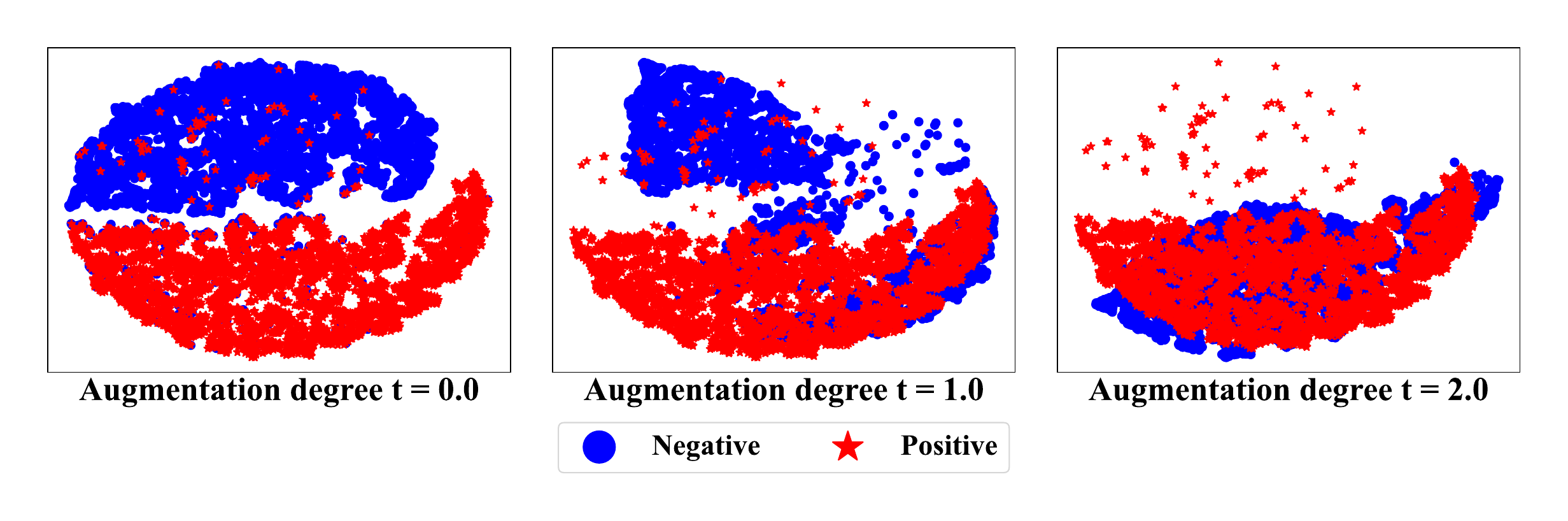}
        \vspace{-20pt}
        \caption{Representations gradually move from negative to positive.}
    \end{subfigure}
    \begin{subfigure}[b]{0.48\linewidth}
        \includegraphics[width=\linewidth]{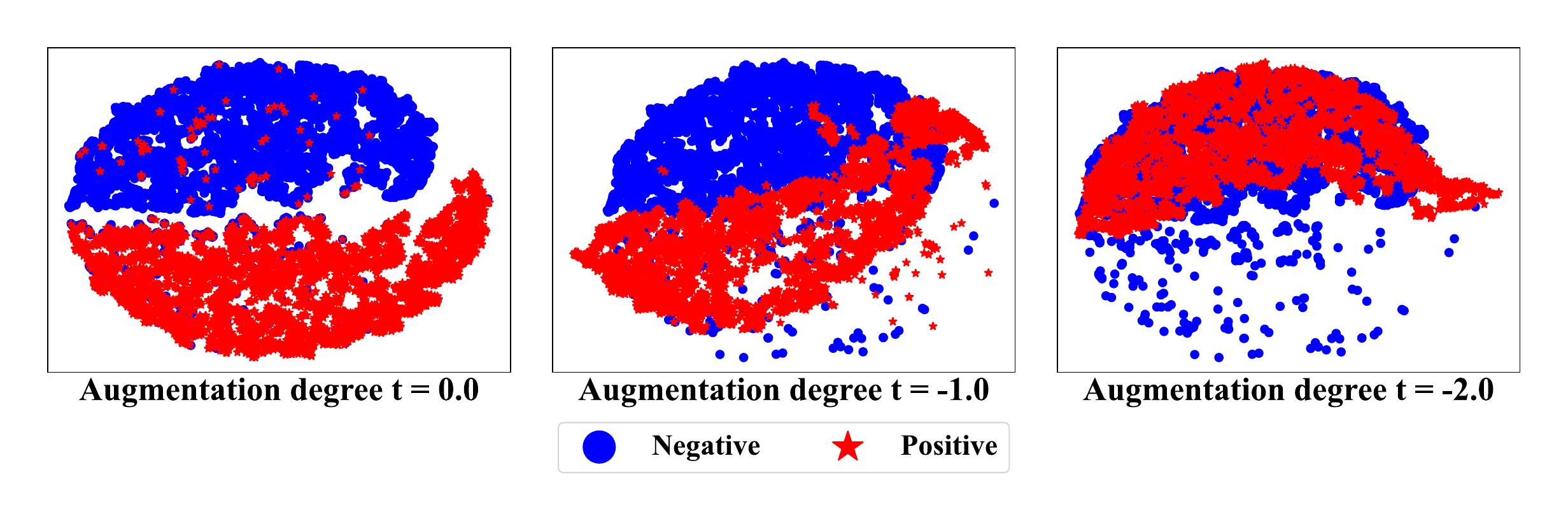}
        \vspace{-20pt}
        \caption{Representations gradually move from positive to negative.}
    \end{subfigure}
    \vspace{-10pt}
    \caption{The effect of augmentation degree $t$ to node representations on pokec\_z.}
    \vspace{-8pt}
    \label{fig:pokecz-aug-tsne}
\end{figure*}
\begin{figure*}[htbp]
    \begin{subfigure}[b]{0.23\linewidth}
        \includegraphics[width=\linewidth]{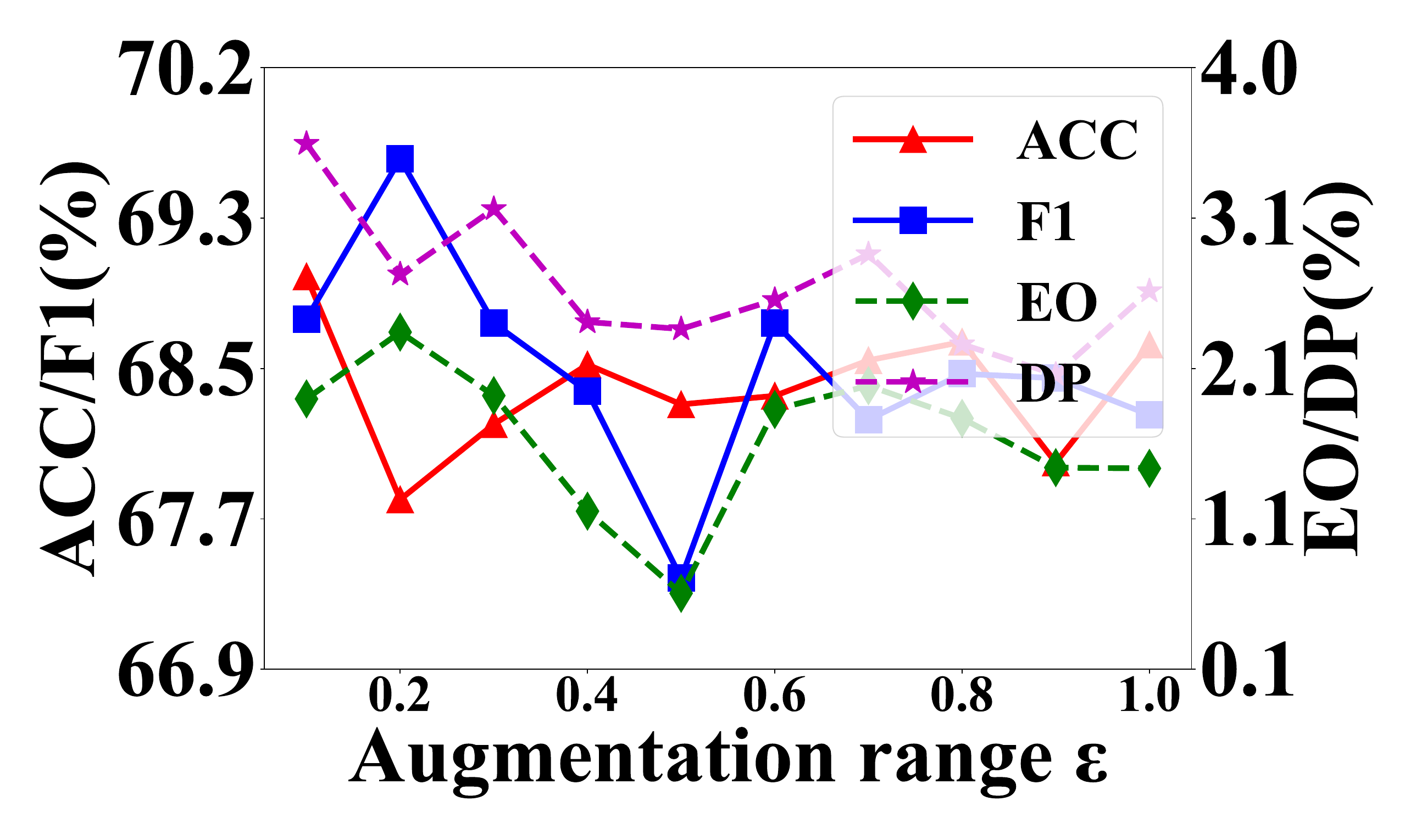}
        \vspace{-15pt}
        \caption{DGI-MinMax.}
    \end{subfigure}
    \begin{subfigure}[b]{0.23\linewidth}
        \includegraphics[width=\linewidth]{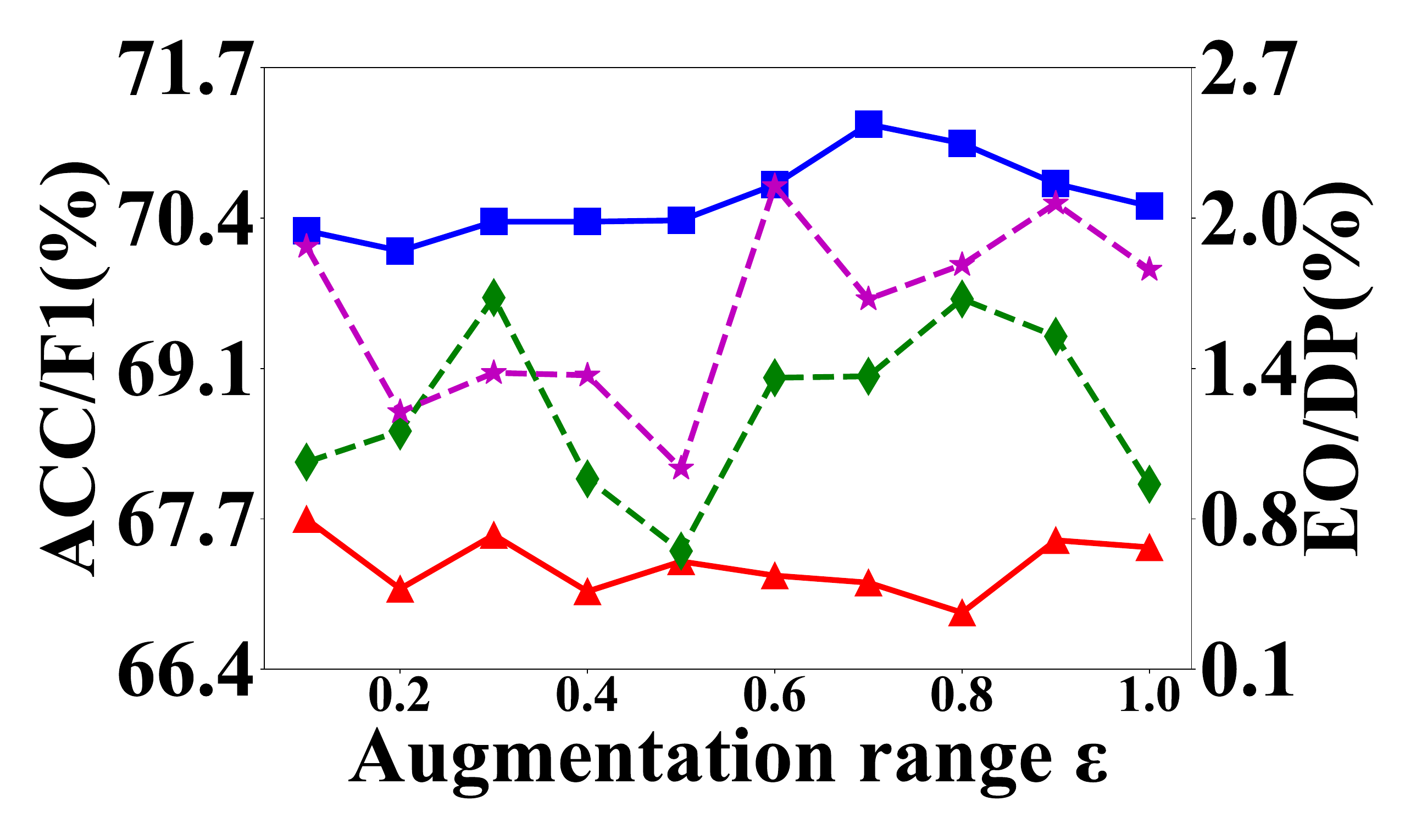}
        \vspace{-15pt}
        \caption{DGI-RandAT.}
    \end{subfigure}
    \begin{subfigure}[b]{0.23\linewidth}
        \includegraphics[width=\linewidth]{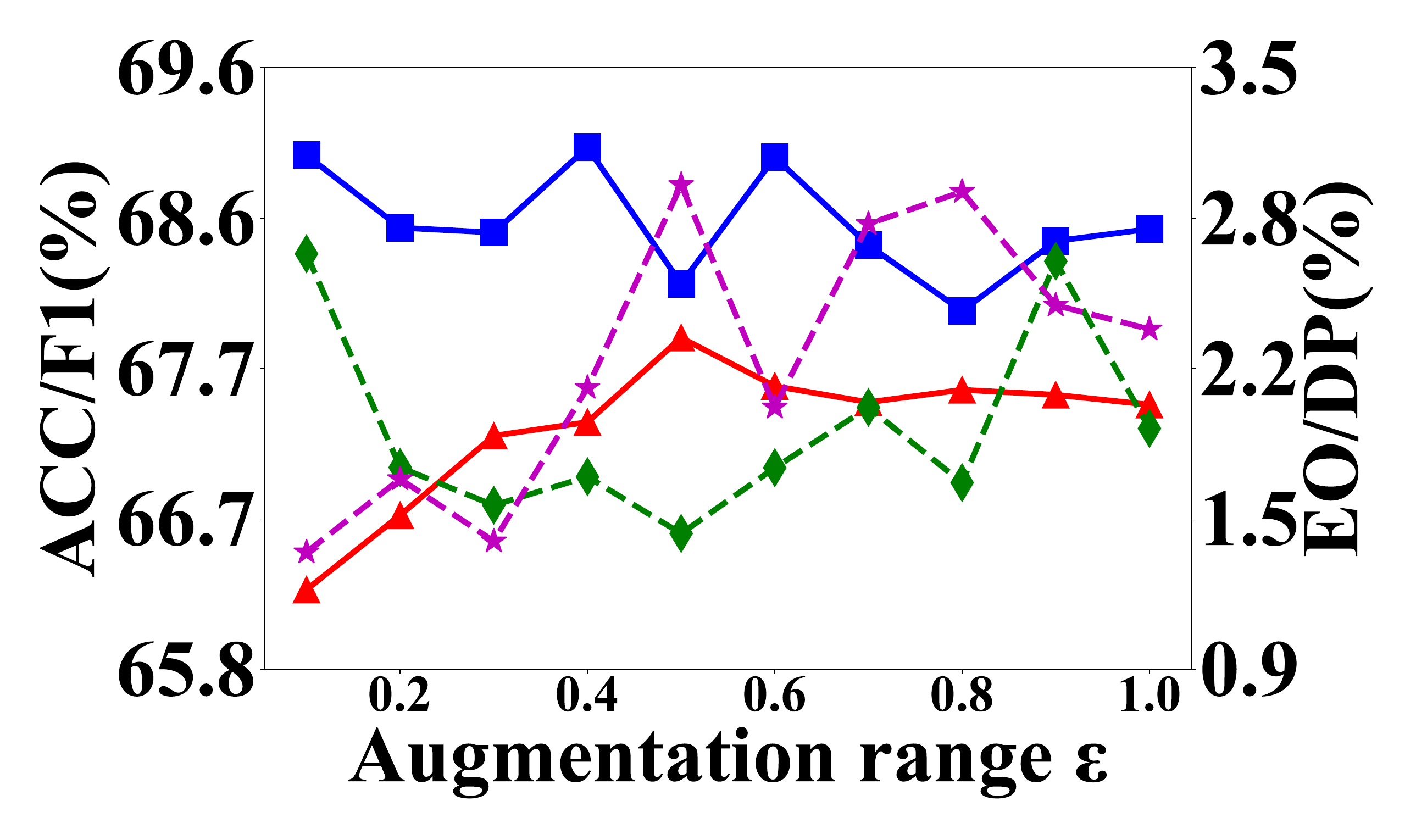}
        \vspace{-15pt}
        \caption{EdgePred-MinMax.}
    \end{subfigure}
    \begin{subfigure}[b]{0.23\linewidth}
        \includegraphics[width=\linewidth]{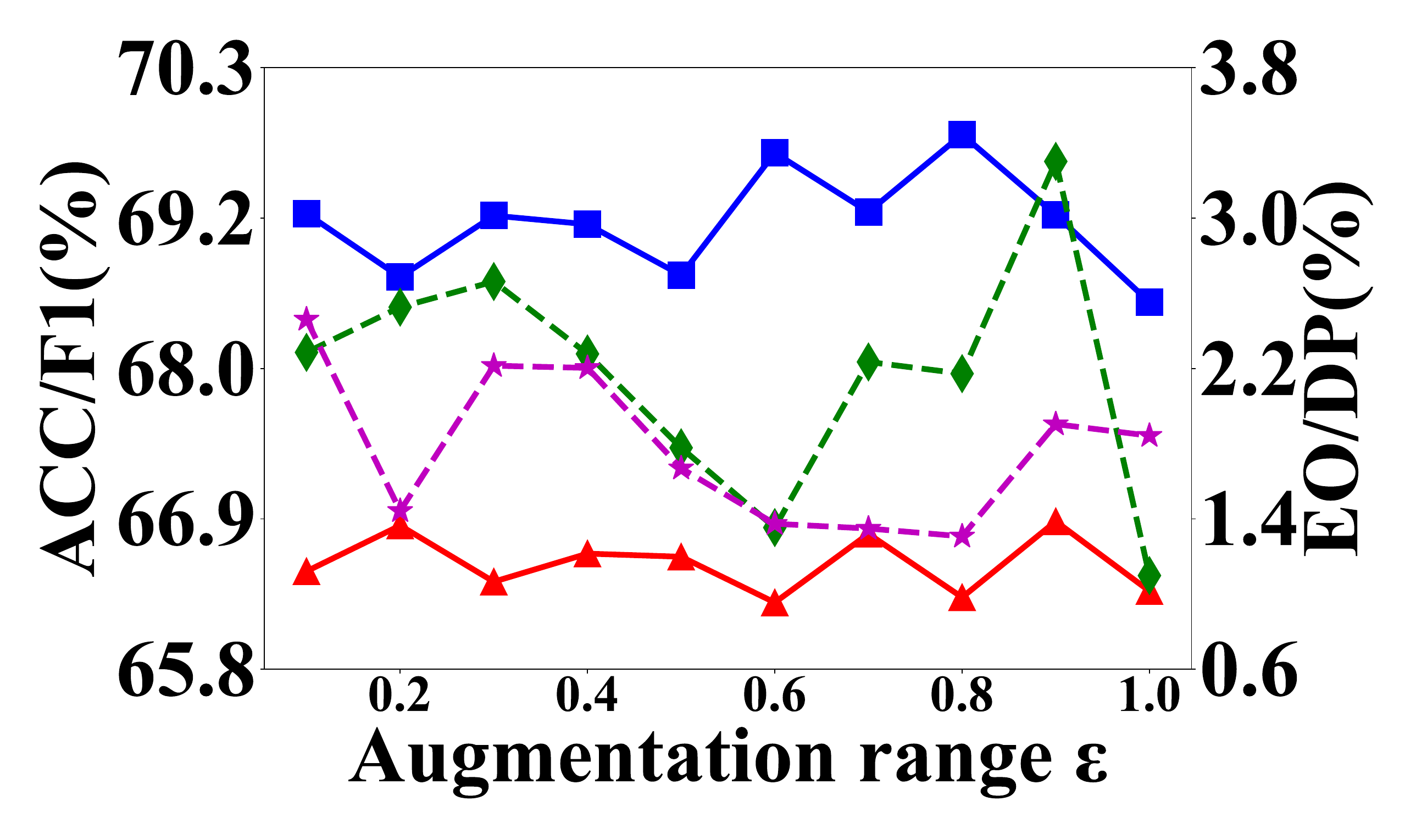}
        \vspace{-15pt}
        \caption{EdgePred-RandAT.}
    \end{subfigure}
    \vspace{-10pt}
    \caption{The effect of augmentation range $\epsilon$ to $\text{GraphPAR}_{minmax}$ and $\text{GraphPAR}_{RandAT}$ on Pokec\_z.}
    \label{fig:sen-pok}
    \vspace{-8pt}
\end{figure*}

\subsection{The Effectiveness of \texorpdfstring{$\boldsymbol \alpha$}. (RQ3)} 
To verify whether $\boldsymbol{\alpha}$ satisfies our expectations, i.e., moving in the direction of $\boldsymbol{\alpha}$ increases the presence of the sensitive attribute while moving in the opposite direction diminishes its presence. 
First, we divide $\mathbf{H}$ into training and test sets. We take the training set to train a sensitive attribute classifier $d_{sens}$ and compute $\boldsymbol{\alpha}$ by Equation~\ref{eq: vector}. 
Next, we randomly construct 100 vectors at angles of 30, 60, and 90 to $\boldsymbol{\alpha}$, respectively. We use $\boldsymbol{\alpha^{\prime}}$ to denote the vector with different angles and the same size as $\boldsymbol{\alpha}$. 
Then, we move the node representations in the test set along the direction of $\boldsymbol{\alpha}$ and $\boldsymbol{\alpha^{\prime}}$ with varying augmentation degree $t$. 
Lastly, we utilize the classifier $d_{sens}$ to predict the accuracy of the sensitive attribute on the test set. We report the average accuracy at different angles, and the results are presented in Figure ~\ref{fig:sens_acc}, revealing the following findings:\\
$\bullet$ When no movement is performed, i.e., $t = 0$, the accuracy is the highest. This again demonstrates that the pre-training inevitably captures the sensitive attribute information present in the dataset.\\
$\bullet$ For the original $\boldsymbol{\alpha}$, i.e., the angle is 0 (red dotted line), as the augmentation degree $|t|$ increases, the prediction accuracy of $d_{sens}$ gradually decreases until it reaches 50\%. This is because modifying the node representations along the same sensitive semantics direction makes all nodes increasingly similar in sensitive attribute semantics. For instance, when moving toward $t>0$, nodes initially classified as negative samples move to positive samples, while nodes previously classified as positive samples remain positive. Consequently, the classifier $d_{sens}$ can only accurately classify half of the nodes. To gain a more concrete understanding, we also visualized the augmentation process using t-SNE. The visualization results are depicted in Figure~\ref{fig:pokecz-aug-tsne}; more results refer to the Appendix~\ref{exp: more-visual}.\\
$\bullet$ For the $\boldsymbol{\alpha^{\prime}}$ that has different angles and the same size as $\boldsymbol{\alpha}$, we observe that as the angle increased, the impact of augmentation on sensitive attribute semantics decreased, meaning the change in accuracy of the sensitive attribute classifier $d_{sens}$ was smaller. This also validates the effectiveness of the direction of $\boldsymbol{\alpha}$.
\begin{figure}[tbp]
    \begin{subfigure}[b]{0.48\linewidth}
        \includegraphics[width=\linewidth]{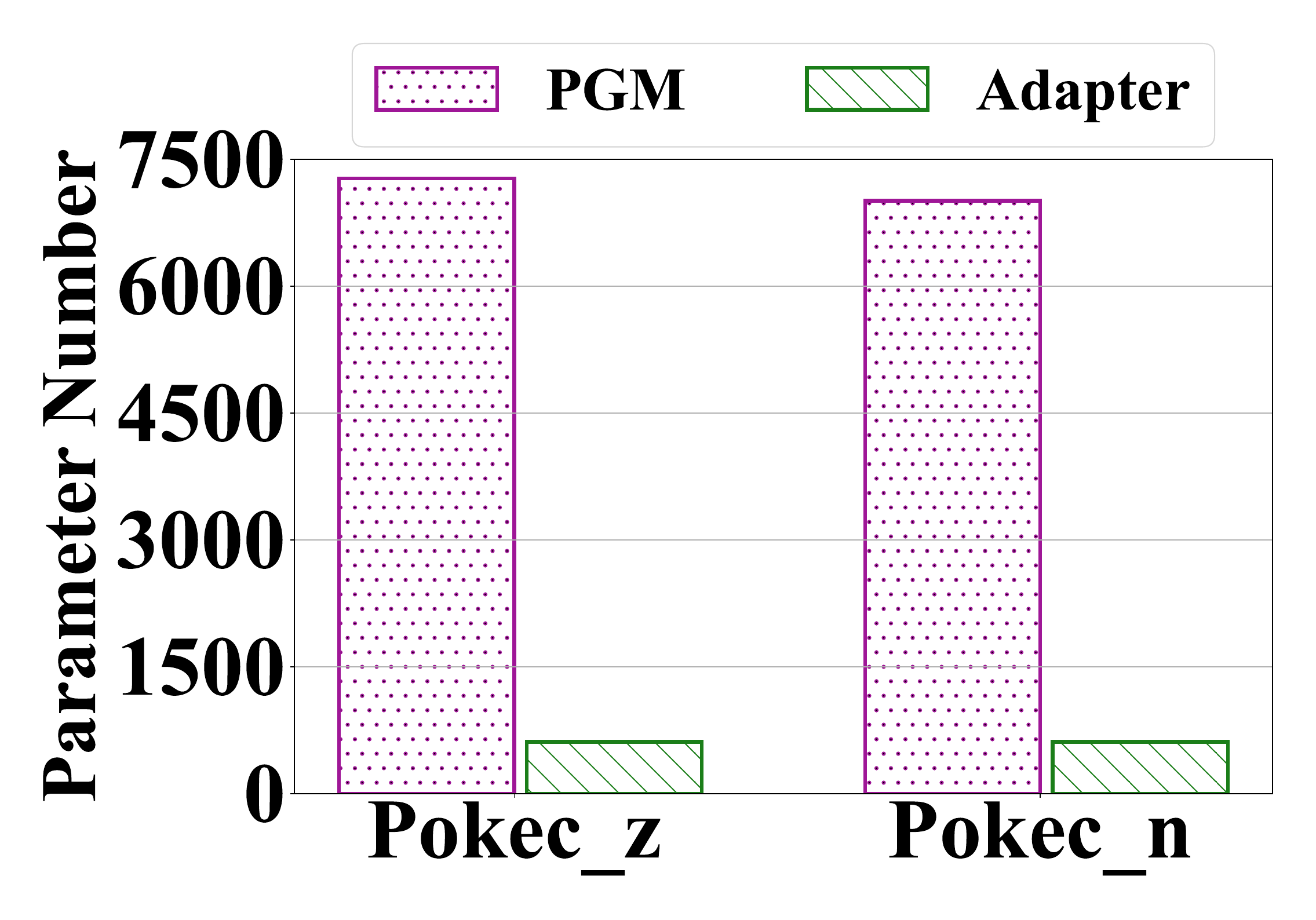}
        \vspace{-15pt}
        \caption{Infomax.}
    \end{subfigure}
    \begin{subfigure}[b]{0.48\linewidth}
        \includegraphics[width=\linewidth]{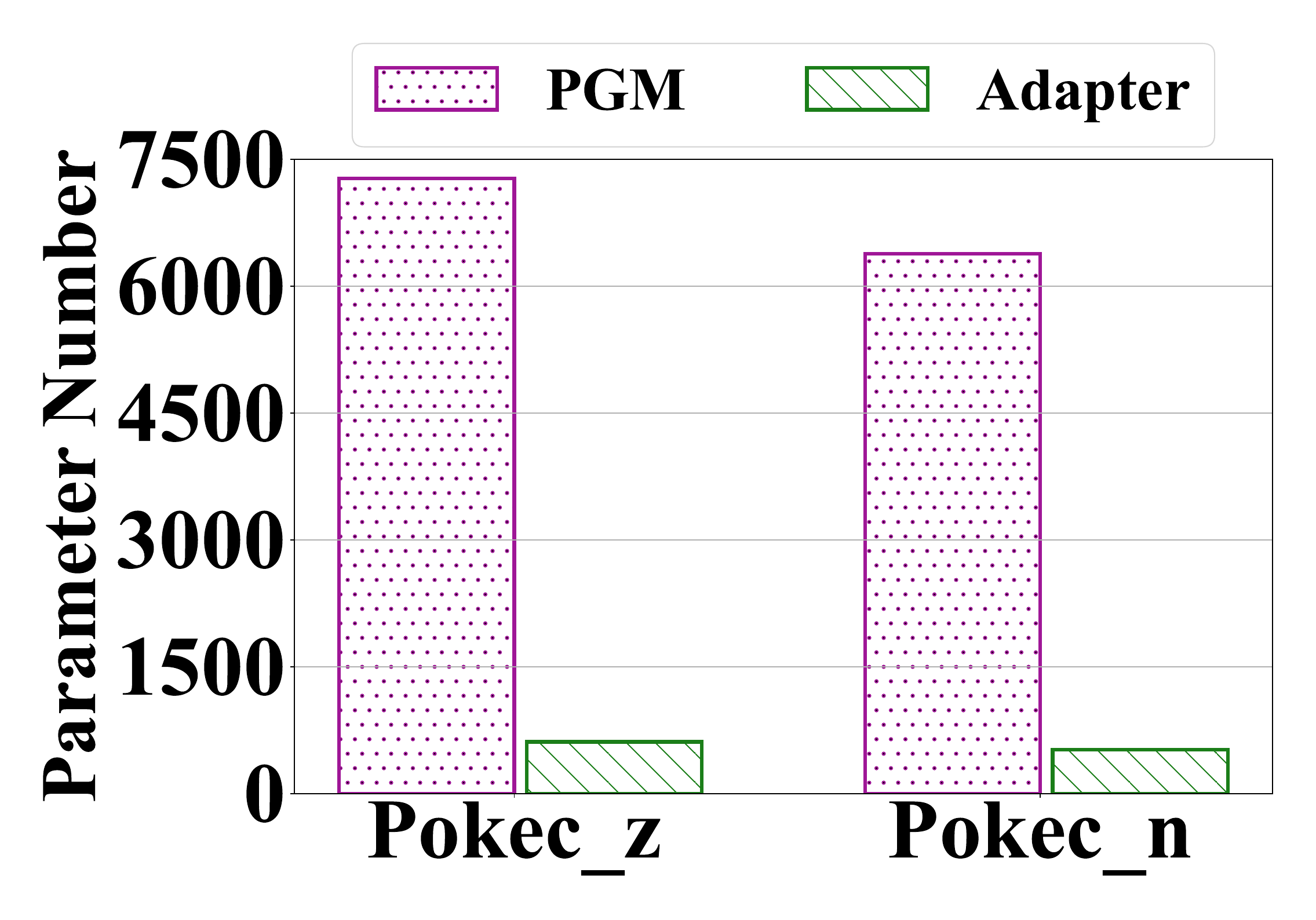}
        \vspace{-15pt}
        \caption{EdgePred.}
    \end{subfigure}
    \vspace{-10pt}
    \caption{Comparison of PGMs and Adapter in the number of parameters tuned.}
    \label{fig:paras}
    \vspace{-15pt}
\end{figure}
\subsection{Hyperparameter Sensitivity Analysis (RQ4)} 
To further validate how the hyperparameters impact the performance of GraphPAR, we conduct hyperparameter sensitivity analysis experiments on the augmentation range $\epsilon$, augmentation sample number $k$, and fairness loss scale $\lambda$. The best hyperparameter $\epsilon, k, \lambda$ for fairness metrics varies among PGMs, datasets, and training methods (RandAT and MinMax). Still, they consistently outperform naive PGMs, illustrating the effectiveness of GraphPAR in improving the fairness of PGMs. For example, as shown in Figure~\ref{fig:sen-pok}, on Pokec\_z trained with MinMax, GraphPAR on DGI achieves the best fairness when $\epsilon = 0.5$, while $\epsilon = 0.3$ for EdgePred. A key observation is that when $\epsilon$ is tuned between 0 and 1, ACC and F1 are stable, while EO and DP fluctuate. This suggests the sensitive semantic augmenter does not corrupt task-related information while successfully capturing sensitive attribute information. More experiment results and analysis refer to the Appendix \ref{App:hyperparameter}.

\subsection{Efficiency Analysis (RQ5)} 
We demonstrate the parameter efficiency of GraphPAR by comparing the parameters in PGMs with the adapter. As shown in Figure~\ref{fig:paras}, the number of tuned parameters in GrahpPAR is 91\% smaller than in the PGM. By contrast, since the parameter of the GNN encoder has to be tuned in existing fair methods, the number of tuned parameters would be equal to or even larger than the size of PGMs, far exceeding that in GraphPAR. In conclusion, GraphPAR is super parameter-efficient, which is well-suited for PGMs.

\section{Conclusion}
In this work, we explore fairness in PGMs for the first time. We discover that PGMs inevitably capture sensitive attribute semantics during pre-training, resulting in unfairness in downstream tasks. To address this problem, we propose GraphPAR to efficiently and flexibly endow PGMs with fairness during the adaptation for downstream tasks. Furthermore, with GraphPAR, we provide theoretical guarantees for fairness. Extensive experiments on real-world datasets demonstrate the effectiveness of GraphPAR in achieving fair predictions and providing provable fairness. In the future, we will further explore other trustworthy directions of PGMs.

\begin{acks}
This work is supported in part by the National Natural Science Foundation of China (No. U20B2045, 61772082, 62002029, 62192784, 62172052, U1936104) and Young Elite Scientists Sponsorship Program (No. 2023QNRC001) by CAST.
\end{acks}

\balance 
\bibliography{reference}

\bibliographystyle{ACM-Reference-Format}

\clearpage
\appendix
\section{Algorithm} \label{algo: graphpar}
We present the whole algorithm process of GraphPAR as follows:
\vspace{-5pt}
\begin{algorithm}[htbp]
\caption{GraphPAR}
\KwData{Graph $\mathcal G=(\mathcal V, \mathcal E, \mathbf X)$, pre-trained graph model $f$}
\KwResult{Adapter $g$ and classifier $d$, and the provable fairness of each node}
\BlankLine
\textbf{1. GraphPAR Training:}\\
Compute the sensitive semantic vector $\boldsymbol{\alpha}$ as Eq~\ref{eq: vector}\;
\For{each epoch}{
    Sample the augmentation set $\mathcal{\hat S}_i$ for each node $i$ as Eq~\ref{eq: set}\;
    \eIf{Train with RandAT}{
    Compute $\mathcal L$ by Eq~\ref{eq: dataaug}\;
    }{
    Compute $\mathcal L$ by Eq~\ref{eq: minmax}\;
    }
    Backward pass with $\mathcal L$\;
}
\BlankLine
\textbf{2. Provide Provable Fairness with Smoothing:}\\
Do adversarial training on the classifier $d$\;
Construct the smoothed adapter $\widehat g$ by Eq~\ref{eq:def-center} and the smoothed classifier $\widehat d$ by Eq~\ref{eq:def-random}\;
\For{each node $i$ in $\mathcal V$}{
Compute the guarantee $d_{cs, i}$ of the adapter as Theorem~\ref{th: center}\;
Compute the guarantee $d_{rs, i}$ of the classifier as Theorem~\ref{th: random}\;
If $d_{cs,i} < d_{rs,i}$, then node $i$ has a provable fairness\;
}
\end{algorithm}
\vspace{-10pt}

\section{Proof of Theorem~\ref{th: provable-fairness}}
\begin{proof}
\label{proof: provable-fairness}
Recall the definition of $g_{\mathbf h}(t) := g(\mathbf h + t \cdot \boldsymbol{\alpha})$ and note that for $\mathbf{h^{\prime}} = \mathbf{h} + t^{\prime} \cdot \boldsymbol{\alpha}$, the center smoothing of 
    \begin{align*}
        \widehat{g_{\mathbf{h}^{\prime}}} (t) & \sim g_{\mathbf{h^{\prime}}} (t + \mathcal{N} (0, \sigma_{cs}^2)) = g (\mathbf{h^{\prime}} + (t + \mathcal{N} (0, \sigma_{cs}^2)) \cdot \boldsymbol{\alpha}), \\
        \widehat{g_{\mathbf{h}}}(t+t^{\prime}) & \sim g_{\mathbf{h}}(t+t^{\prime}+\mathcal{N} (0, \sigma_{cs}^2))=g (\mathbf h + (t + t^{\prime} + \mathcal{N} (0, \sigma_{cs}^2)) \cdot \boldsymbol{\alpha}).
    \end{align*}    
    
    Since $\mathbf{h^{\prime}} = \mathbf{h} + t^{\prime} \cdot \boldsymbol{\alpha}$, the sampling distributions are the same, hence $\widehat{g_{\mathbf{h^{\prime}}}} (t) = \widehat{g_{\mathbf h}}(t+t^{\prime})$, and in particular $\widehat{g}(\mathbf h^{\prime})=\widehat{g_{\mathbf h^{\prime}}}(0)=\widehat{g_{\mathbf h}}(t^{\prime})$. 
    
    Now, let us get back to Equation~\ref{eq: def}. By definition of $\mathcal S$, for all $\mathbf{h^{\prime}} \in \mathcal S$, $\mathbf{h^{\prime}} = \mathbf{h} + t^{\prime} \cdot \boldsymbol{\alpha}$ for some $t^{\prime} \in [-\epsilon, \epsilon]$. Moreover, $\mathbf{z}_{cs} = \widehat{g}(\mathbf{h})=\widehat{g_{\mathbf{h}}}(0)$ and $\widehat{g}(\mathbf{h^{\prime}})=\widehat{g_{\mathbf{h}}}(t^{\prime})$. Theorem~\ref{th: center} tells us that with confidence $1 - \alpha_{cs}$:
    \begin{align*}     
        &\left\|\widehat{g_{\mathbf{h}}}\left(0\right)-\widehat{g_{\mathbf{h}}}\left(t^{\prime}\right)\right\|_{2}\leq d_{cs},\ \forall \ t^{\prime}\in\left[-\epsilon,\epsilon\right] \\
        \Longleftrightarrow &\left\|\mathbf{z}_{cs}-\widehat{g}\left(\mathbf{h}^{\prime}\right)\right\|_{2}\leq d_{cs},\ \forall \ \mathbf{h}^{\prime}\in \mathcal S, \numberthis \label{eq: center}
    \end{align*}
    provided that the center smoothing computation of $d_{cs}$ does not abstain.
    
    Finally, we consider the last component of the pipeline, i.e., the smoothed classifier $\widehat d$. Provided that $\widehat d$ does not abstain at the input $d_{cs}$, Theorem~\ref{th: random} provides us with a radius $d_{rs}$ around $\mathbf{z}_{cs}$ such that with confidence $1 - \alpha_{rs}$:
\begin{align*}
& \widehat{d}\left(\mathbf{z}_{cs}\right)=\widehat{d}\left(\mathbf{z}_{cs}+\boldsymbol{\delta}\right),\ \forall\boldsymbol{\delta}\mathrm{~s.t.~}\|\boldsymbol{\delta}\|_2<d_{rs} \\
\Longleftrightarrow & \widehat{d}\left(\mathbf{z}_{cs}\right) =\widehat{d}\left(\mathbf{z}^{\prime}\right),\ \forall\mathbf{z}^{\prime}\mathrm{~s.t.~}\left\|\boldsymbol{z}_{cs}-\mathbf{z}^{\prime}\right\|_{2}<d_{rs}.  \numberthis \label{eq: random}
\end{align*}

    If $d_{cs} < d_{rs}$, combining the conclusions in Equation~\ref{eq: center} and Equation~\ref{eq: random} and applying the union bound, we obtain that with confidence $1 - \alpha_{cs} - \alpha_{rs}$ we have $\widehat{d}(\mathbf{z}_{cs}) = \widehat{d}(\widehat{g}(\mathbf{h}^{\prime}))$ for all $\mathbf{h}^{\prime} \in \mathcal{S}$, that is,
    \begin{equation}
    \forall \ \mathbf{h}^{\prime}\in \mathcal{S}\left(\mathbf{h}\right):\widehat{d}\circ\widehat{g}\left(\mathbf{h}\right)=\widehat{d}\circ\widehat{g}\left(\mathbf{h}^{\prime}\right)
\end{equation}
as required by Definition~\ref{def: provable-fair}. The same proof technique can extend to the multiple sensitive attribute vectors case. 
\end{proof} 

\section{More Experimental Details}
\subsection{Datasets}
\label{exp: detail-dataset}
Detailedly, \textit{Income}~\cite{asuncion2007uci} is collected from the Adult Data Set. The sensitive attribute is race, and the task is to classify whether an individual salary exceeds 50,000\$. 
\textit{Credit}~\cite{agarwal2021towards} encompasses a network of individuals connected according to the likeness of their spending and payment habits. The sensitive attribute is the age of these individuals, and the objective is to predict whether their default payment method is credit card or not.
\textit{Pokec\_z} and \textit{Pokec\_n}~\cite{dai2021say} are created by sampling from Pokec based on geographic regions. Pokec encompasses anonymized data from the complete social network in 2012. The sensitive attribute is the region, and the predicted label is the working field. 

\subsection{Implementations}
\label{exp: exp-set}
Unless otherwise specified, we set the hyperparameters as follows: For the sensitive semantics augmented, sensitive attribute semantics augmentation range $\epsilon=0.5$, number of randomly selected augmentation samples $k=20$, fairness loss scale factor $\lambda=0.1$. For the adapter, the dimension size of the down projection is half of the input, the learning rate is 0.01, and the training epoch is 1000. We use GCN as the backbone for all PGMs and take the Adam optimizer. Referring to random smooth~\cite{cohen2019certified}, after adapting PGMs to downstream tasks, we uniformly utilize Gaussian data augmentation with a variance of 1 to additionally adversarial train the classifier for 100 rounds, which maximizes the number of nodes with provable fairness without compromising accuracy. Following the parameter settings in center smooth~\cite{kumar2021center} and random smooth~\cite{cohen2019certified}, we utilize the well-trained adapter and the classifier to construct their smoothed versions, respectively. In the future, we will also provide an implementation based on GammaGL~\cite{liu2023gammagl} at \href{https://github.com/BUPT-GAMMA/GammaGL}{https://github.com/BUPT-GAMMA/GammaGL}.
\begin{figure*}[htbp]
    \begin{subfigure}[b]{0.48\linewidth}
        \includegraphics[width=\linewidth]{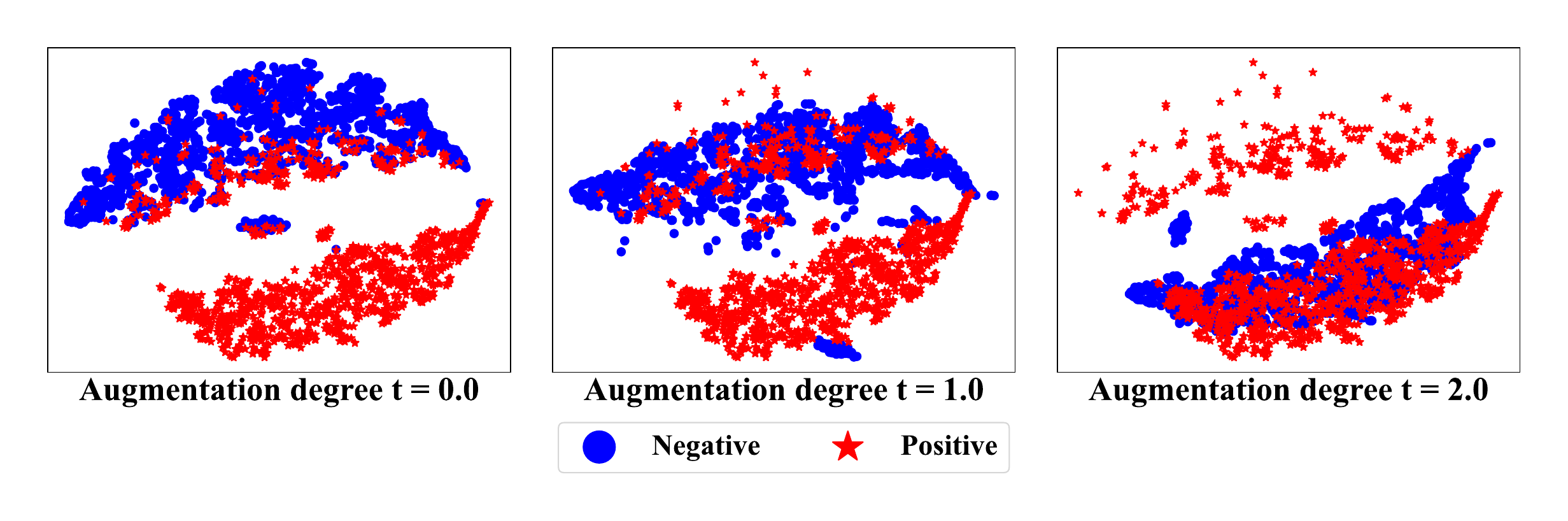}
        \vspace{-20pt}
        \caption{Representations gradually move from negative to positive.}
    \end{subfigure}
    \begin{subfigure}[b]{0.48\linewidth}
        \includegraphics[width=\linewidth]{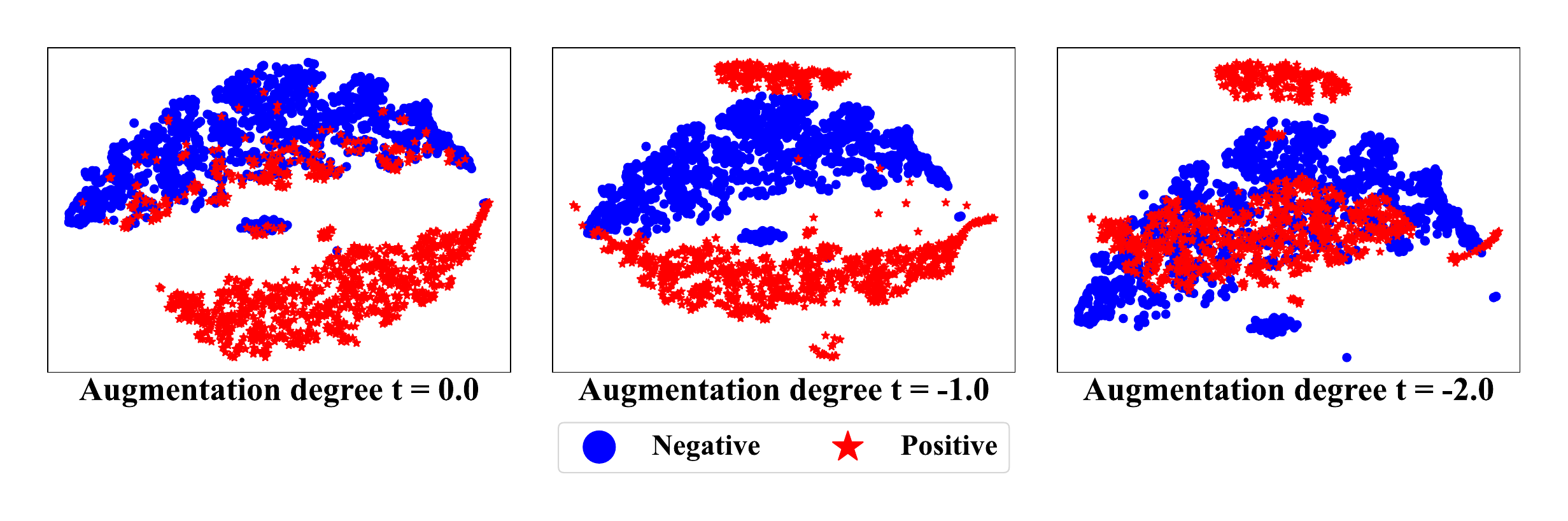}
        \vspace{-20pt}
        \caption{Representations gradually move from positive to negative.}
    \end{subfigure}
    \vspace{-10pt}
    \caption{The effect of augmentation degree $t$ to node representations on Income.}
    \vspace{-5pt}
    \label{fig:income-aug-tsne}
\end{figure*}

\begin{figure*}[htbp]
    \begin{subfigure}[b]{0.48\linewidth}
        \includegraphics[width=\linewidth]{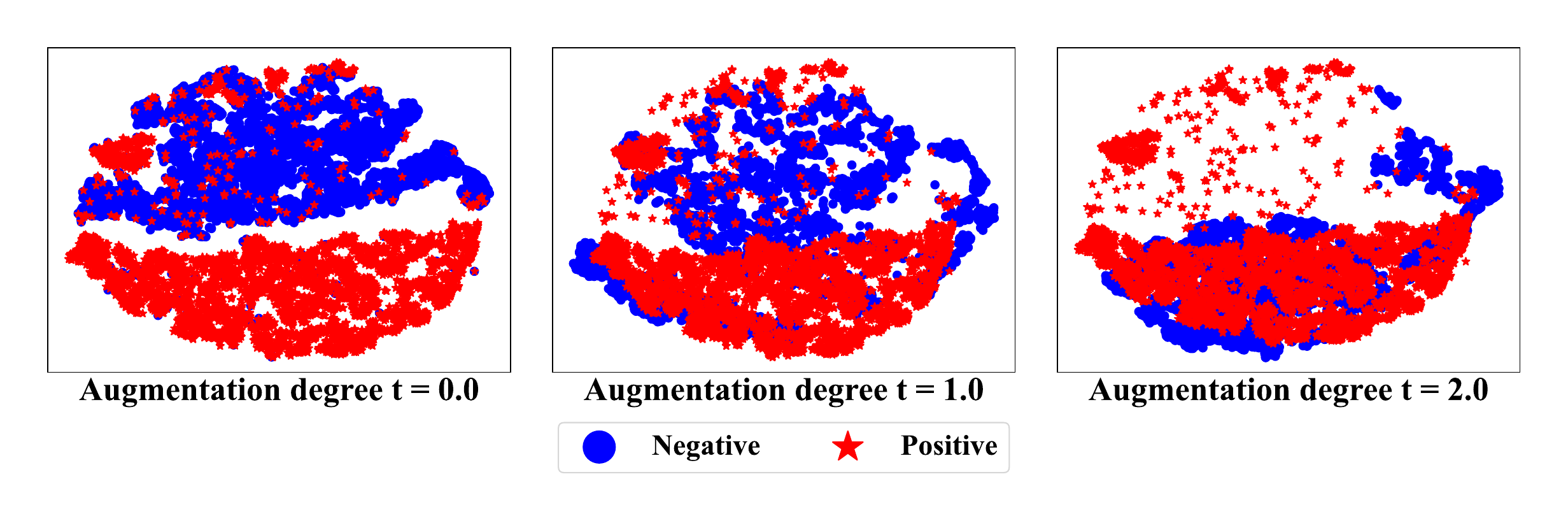}
        \vspace{-20pt}
        \caption{Representations gradually move from negative to positive.}
    \end{subfigure}
    \begin{subfigure}[b]{0.48\linewidth}
        \includegraphics[width=\linewidth]{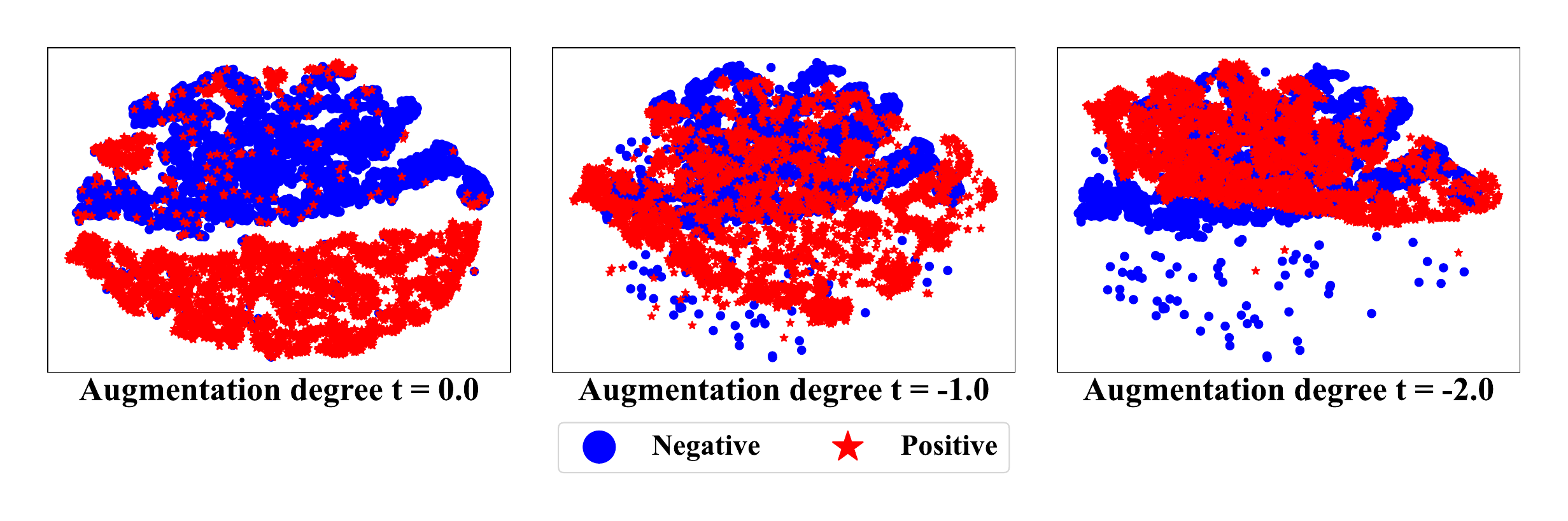}
        \vspace{-20pt}
        \caption{Representations gradually move from positive to negative.}
    \end{subfigure}
    \vspace{-10pt}
    \caption{The effect of augmentation degree $t$ to node representations on pokec\_n.}
    \vspace{-5pt}
    \label{fig:pokecn-aug-tsne}
\end{figure*}

\subsection{Baselines}
\label{exp: detail-baseline}
We compare GraphPAR to four baseline models: GCN~\cite{kipf2016semi} is the most common GNN; FairGNN~\cite{dai2021say} is a framework for fair node classification using GNNs given limited sensitive attribute information; NIFTY~\cite{agarwal2021towards} achieves fairness by maximizing the similarity of representations learned from the original graph and their augmented counterfactual graphs. EDITS~\cite{dong2022edits} debiases the input network to remove the sensitive information in the graph data. Since GraphPAR is based on PGMs, we include three types of PGM as baseline models: contrastive pre-training models DGI~\cite{DGI2018} and GCA~\cite{zhu2021graph} that maximize the mutual information between different views, as well as predictive pre-training model EdgePred~\cite{hamilton2017inductive} that reconstructs masked edges as its task. 

\begin{table}[htbp]
 \centering
 \vspace{3pt}
 \caption{Performance and fairness ($\%\pm\sigma$) on node classification. The best results are in bold and runner-up results are underlined.} 
 \vspace{-5pt}
 \label{tab:income-node-classification}
 \begin{adjustbox}{width=0.45\textwidth}
 \Huge
\begin{tabular}{cccccc}
\hline
\toprule
\multicolumn{2}{c}{\multirow{2}{*}{Method}}            & \multicolumn{4}{c}{Income}                                                          \\ \cline{3-6} 
\multicolumn{2}{c}{}                                   & ACC (↑)             & F1 (↑)              & DP (↓)             & EO (↓)             \\ \hline
\multicolumn{2}{c}{GCN}                                & 69.08±0.35          & \textbf{49.39±0.13} & 29.73±1.43         & 33.54±3.43         \\
\multicolumn{2}{c}{FairGNN}                            & 68.90±1.49          & 47.26±0.70          & 15.39±2.45         & 21.51±3.59         \\
\multicolumn{2}{c}{NIFTY}                              & 70.37±1.86          & 47.87±0.33          & 26.84±1.27         & 29.09±1.53         \\
\multicolumn{2}{c}{EDITS}                              & 69.02±0.59          & {\ul 49.21±0.37}    & 27.11±2.76         & 31.11±4.23         \\ \hline
\multirow{3}{*}{DGI}      & Naive                      & {\ul 76.62±0.60}    & 48.15±2.35          & 23.07±5.81         & 30.26±7.59         \\
                          & $\text{GraphPAR}_{RandAT}$ & \textbf{76.63±1.10} & 46.94±1.19          & 15.43±4.48         & 19.80±7.93         \\
                          & $\text{GraphPAR}_{MinMax}$ & 75.29±1.60          & 47.27±1.08          & \textbf{8.75±1.33} & \textbf{7.90±3.84} \\ \hline
\multirow{3}{*}{EdgePred} & Naive                      & 69.15±2.02          & 46.34±2.63          & 29.73±3.19         & 35.79±7.83         \\
                          & $\text{GraphPAR}_{RandAT}$ & 70.41±2.03          & 45.51±3.27          & 23.51±7.42         & 28.40±13.50        \\
                          & $\text{GraphPAR}_{MinMax}$ & 69.06±3.69          & 46.58±1.28          & 11.68±7.06         & {\ul 14.18±7.65}   \\ \hline
\multirow{3}{*}{GCA}      & Naive                      & 75.00±2.10          & 46.91±3.76          & 21.52±6.25         & 27.73±9.08         \\
                          & $\text{GraphPAR}_{RandAT}$ & 74.95±1.41          & 46.55±2.72          & 16.72±3.80         & 22.30±6.05         \\
                          & $\text{GraphPAR}_{MinMax}$ & 75.44±1.79          & 46.70±1.74          & {\ul 9.92±3.75}    & 16.99±4.56         \\ \hline
\end{tabular}
\end{adjustbox}

\end{table}

\section{Additional Experiment Analysis}
\subsection{Effectiveness of GraphPAR on Income} 
\label{exp: effect-income}
As shown in Table~\ref{tab:income-node-classification} and Table~\ref{tab:income-cert-fair}, similar to performance on the Credit, Pokec\_n, and Pokec\_z datasets, GraphPAR outperforms baseline models in terms of classification performance and fairness. By employing the two proposed adapter tuning methods, GraphPAR significantly enhances the fairness of PGMs in downstream tasks without nearly compromising prediction performance. Moreover, based on $\text{GraphPAR}_{MinMax}$, around 83\% of nodes exhibit provable fairness.

\subsection{More Visualization Results on \texorpdfstring{$\boldsymbol \alpha$}. }
\label{exp: more-visual}
To gain a more concrete understanding of the augmentation process in the direction of $\mathbf{\alpha}$, we also visualized the augmentation process using t-SNE on Income and Pokec\_n datasets, and the visualization results depicted in Figure~\ref{fig:income-aug-tsne} and Figure~\ref{fig:pokecn-aug-tsne}, respectively.

\begin{table}[htbp]
\centering
\vspace{-6pt}
\caption{Provable fairness under different training schemes.}
\vspace{-10pt}
\label{tab:income-cert-fair}
\begin{adjustbox}{width=0.45\textwidth}
\begin{tabular}{cccccccc}
\hline
\toprule
\multirow{2}{*}{\huge{Dataset}}                     & \multirow{2}{*}{\huge{PGM}} & \multicolumn{2}{c}{\huge{Naive}} & \multicolumn{2}{c}{\huge{$\text{GraphPAR}_{RandAT}$}} & \multicolumn{2}{c}{\huge{$\text{GraphPAR}_{MinMax}$}} \\ \cline{3-8} 
                                             &                      & \Large{ACC (↑)}  & \Large{Prov\_Fair (↑)} & \Large{ACC (↑)}                & \Large{Prov\_Fair (↑)}        & \Large{ACC (↑)}            & \Large{Prov\_Fair (↑)}            \\ \hline
\multicolumn{1}{c|}{\multirow{3}{*}{\huge{Income}}} & \Large{DGI}                  & \huge{72.85}    & \huge{0.02 }          & \textbf{\huge{73.42}}         & \huge{0.94 }                 & \huge{73.19}              & \textbf{\huge{81.01}}            \\
\multicolumn{1}{c|}{}                        & \Large{EdgePred}             & \huge{67.17}    & \huge{0.01 }          & \textbf{\huge{68.24}}         & \huge{7.64 }                 & \huge{66.09}              & \textbf{\huge{80.62}}            \\
\multicolumn{1}{c|}{}                        & \Large{GCA}                  & \huge{70.91}    & \huge{0.45 }          & \textbf{\huge{71.59}}         & \huge{5.84 }                 & \huge{72.47}              & \textbf{\huge{90.68}}            \\ \hline
\end{tabular}
\end{adjustbox}
\vspace{-15pt}
\end{table}

\subsection{More Hyperparameter Analysis}\label{App:hyperparameter}
\begin{figure*}[htbp]
    \begin{subfigure}[b]{0.23\linewidth}
        \includegraphics[width=\linewidth]{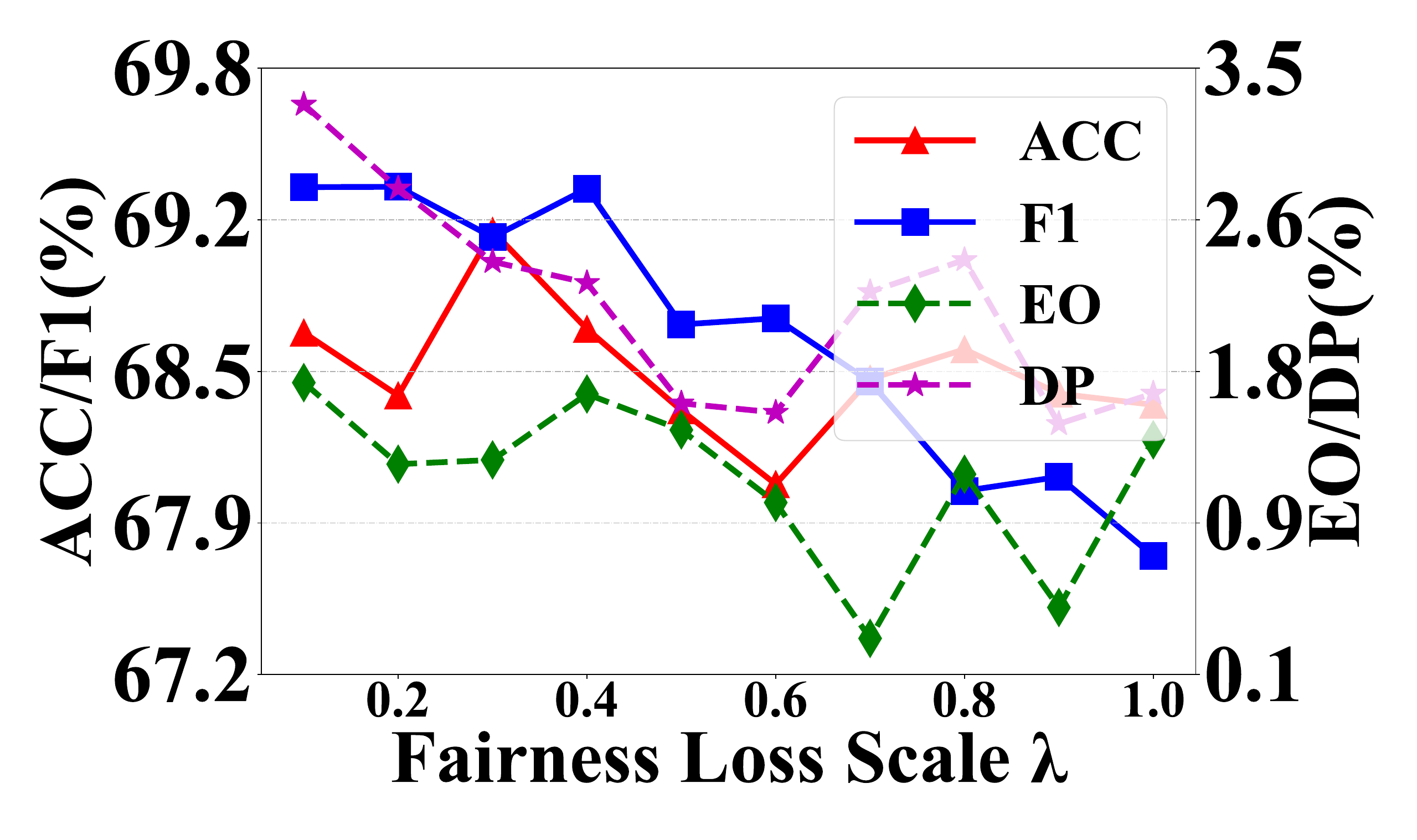}
        \caption{DGI-Pokec\_z.}
    \end{subfigure}
    \begin{subfigure}[b]{0.23\linewidth}
        \includegraphics[width=\linewidth]{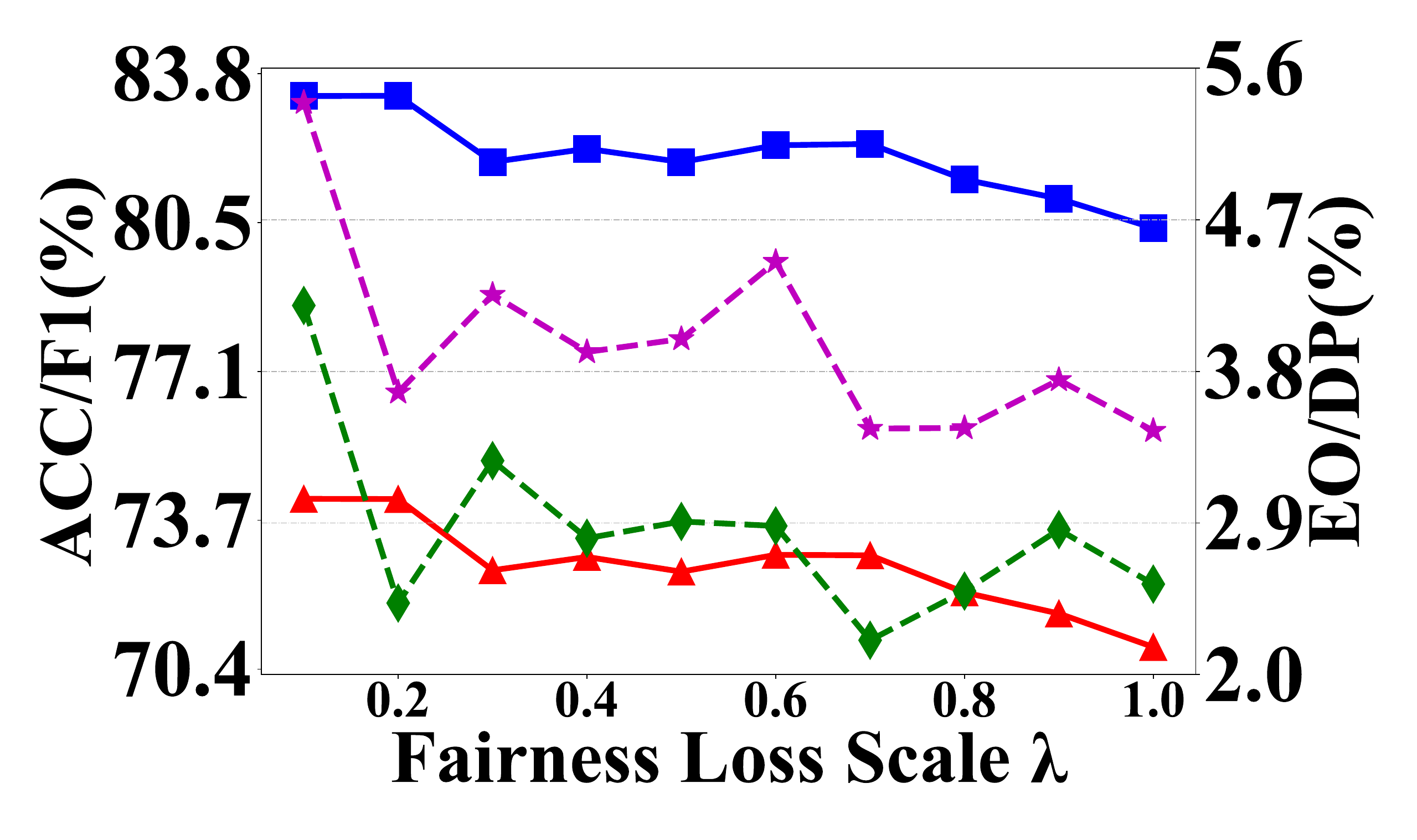}
        \caption{DGI-Credit.}
    \end{subfigure}
    \begin{subfigure}[b]{0.23\linewidth}
        \includegraphics[width=\linewidth]{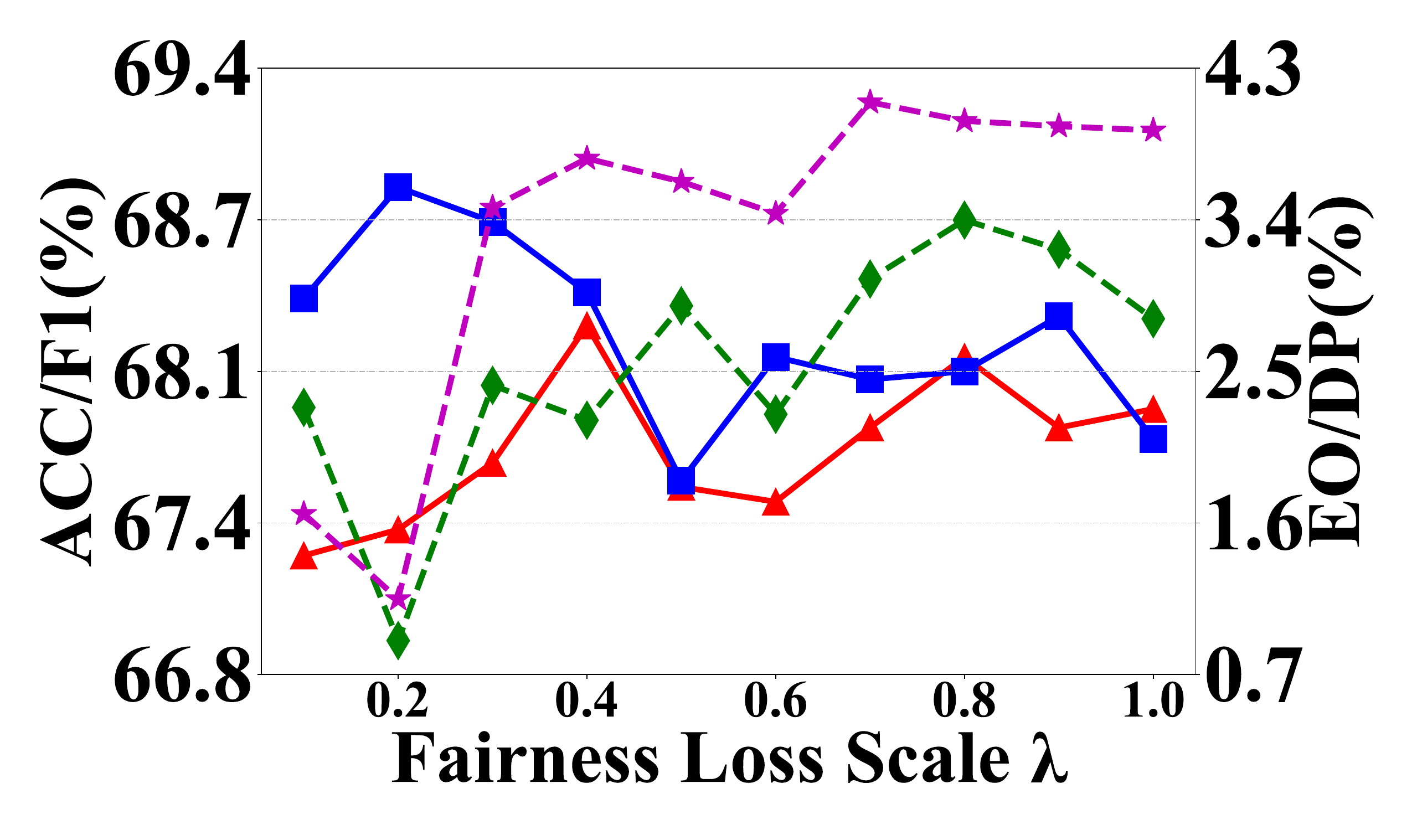}
        \caption{EdgePred-Pokec\_z.}
    \end{subfigure}
    \begin{subfigure}[b]{0.23\linewidth}
        \includegraphics[width=\linewidth]{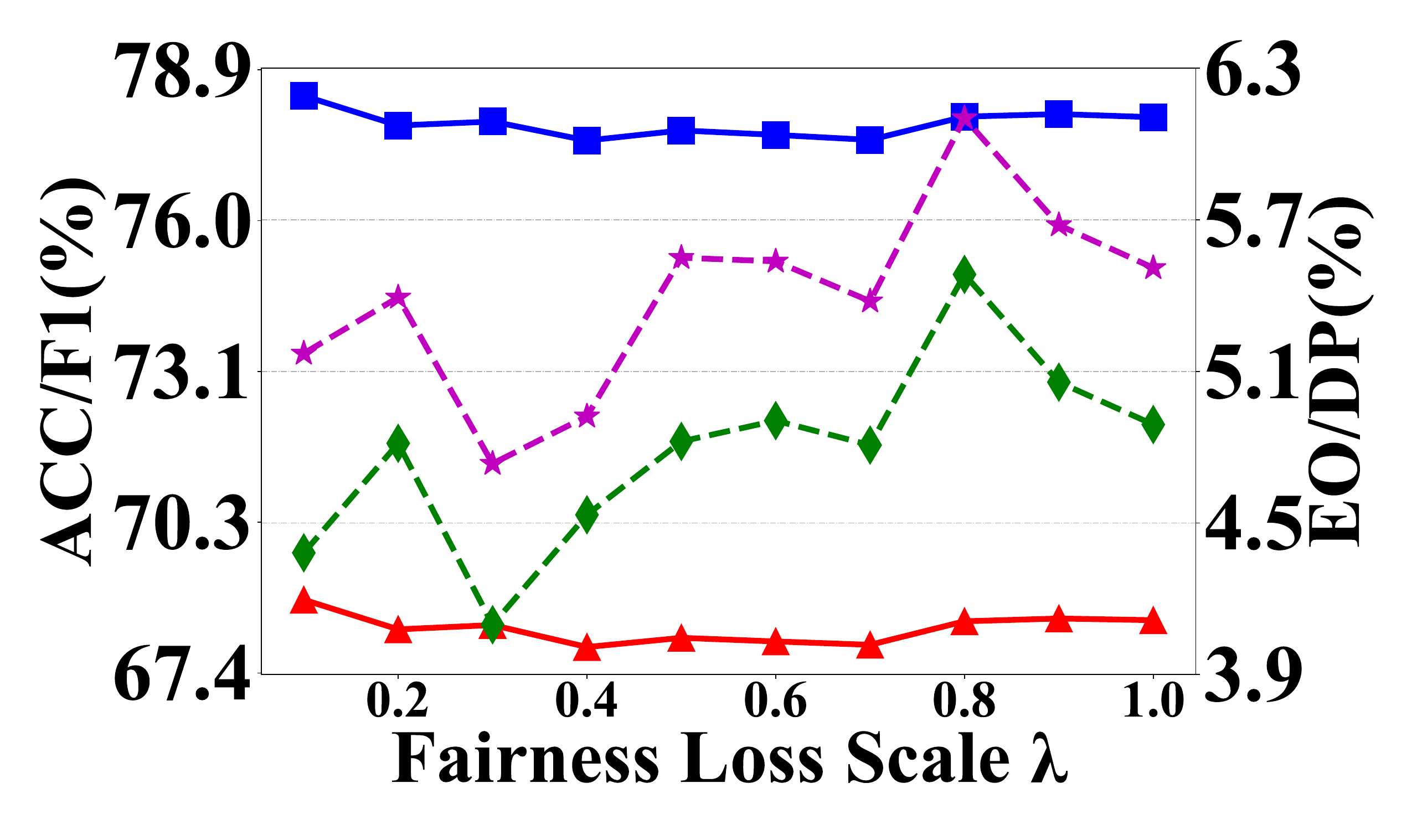}
        \caption{EdgePred-Credit.}
    \end{subfigure}
    \caption{The effect of fairness loss scale factor $\lambda$ to $\text{GraphPAR}_{minmax}$.}
    \label{fig:sen-lamda}
\end{figure*}

\begin{figure*}[htbp]
    \begin{subfigure}[b]{0.23\linewidth}
        \includegraphics[width=\linewidth]{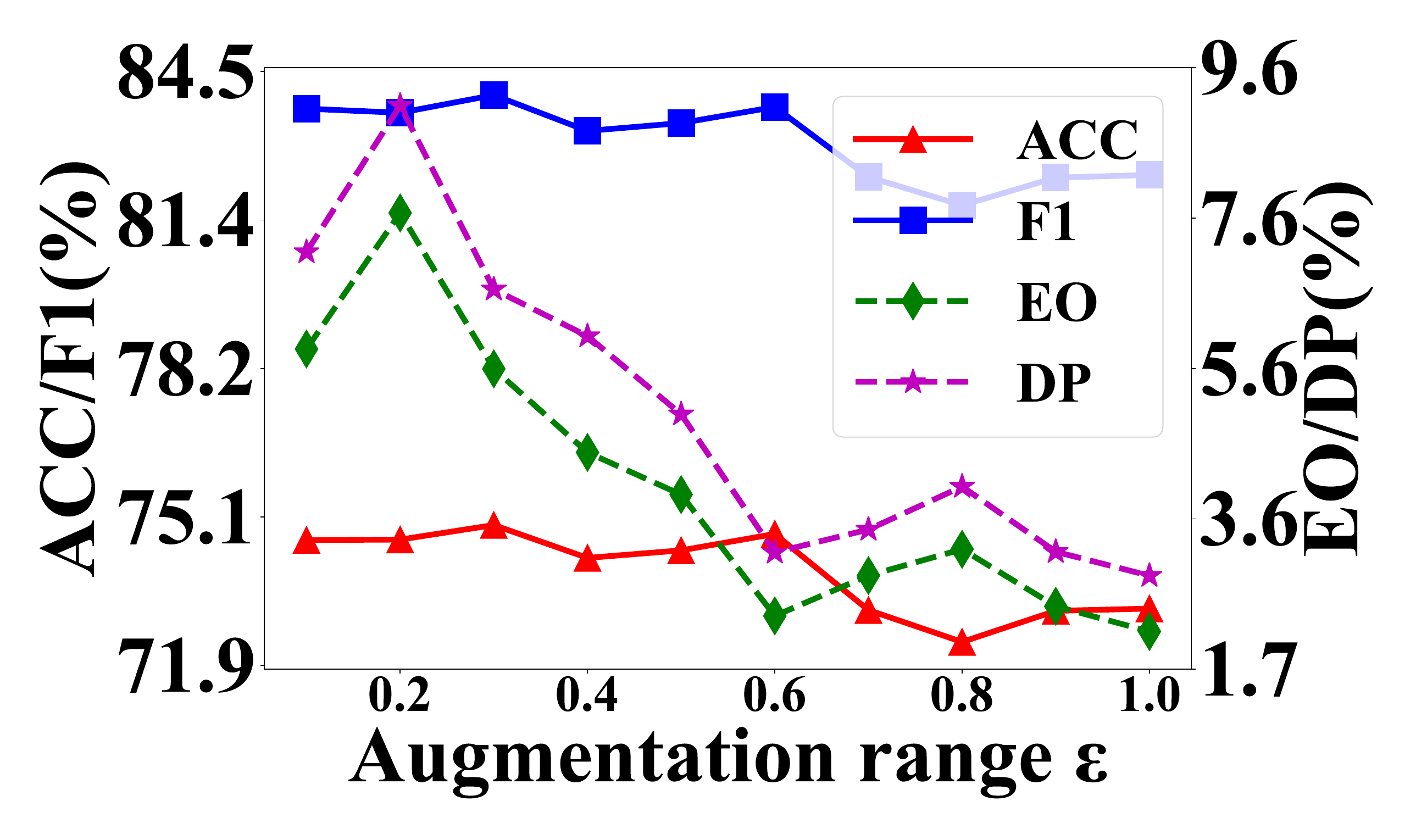}
        \caption{DGI-MinMax.}
    \end{subfigure}
    \begin{subfigure}[b]{0.23\linewidth}
        \includegraphics[width=\linewidth]{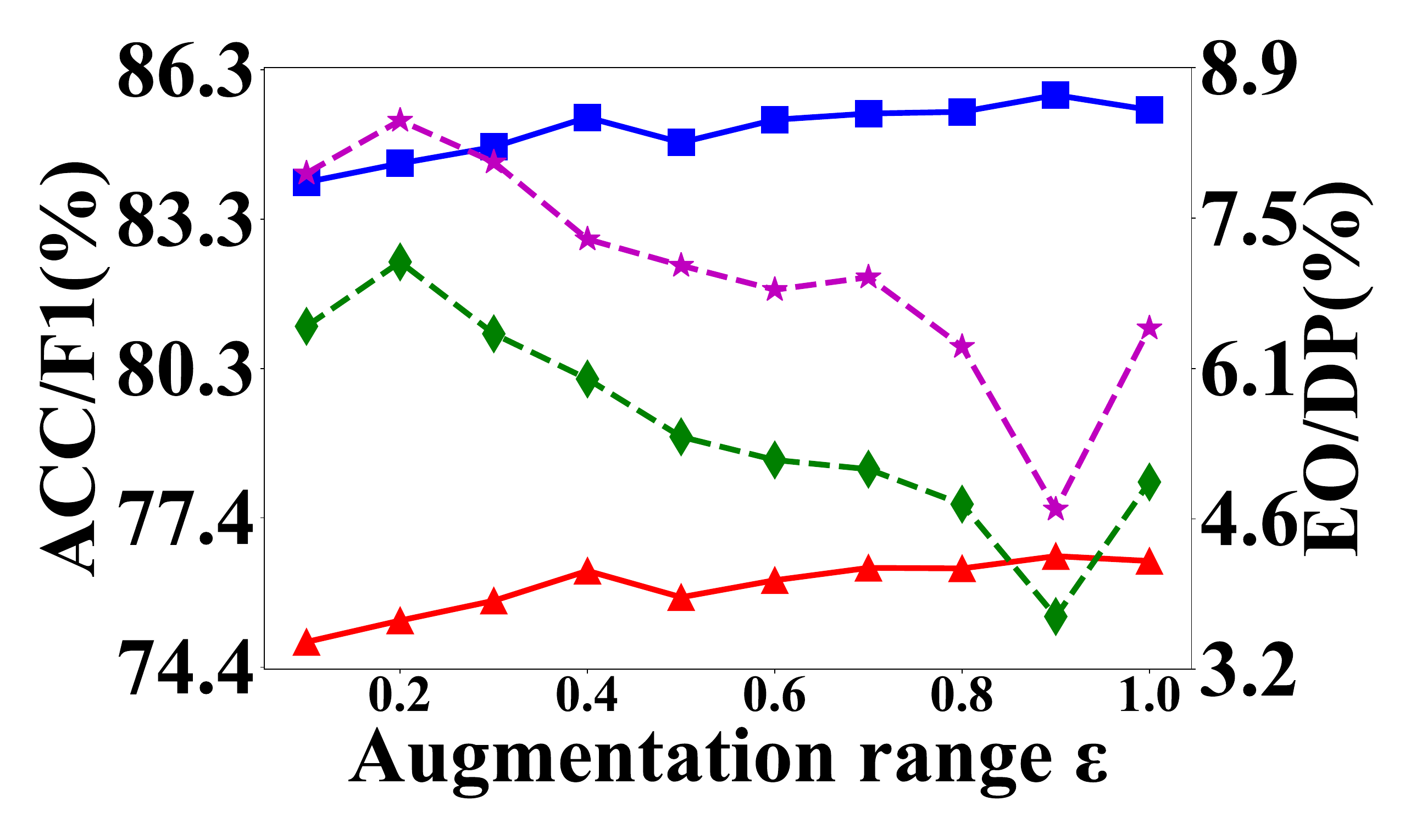}
        \caption{DGI-RandAT.}
    \end{subfigure}
    \begin{subfigure}[b]{0.23\linewidth}
        \includegraphics[width=\linewidth]{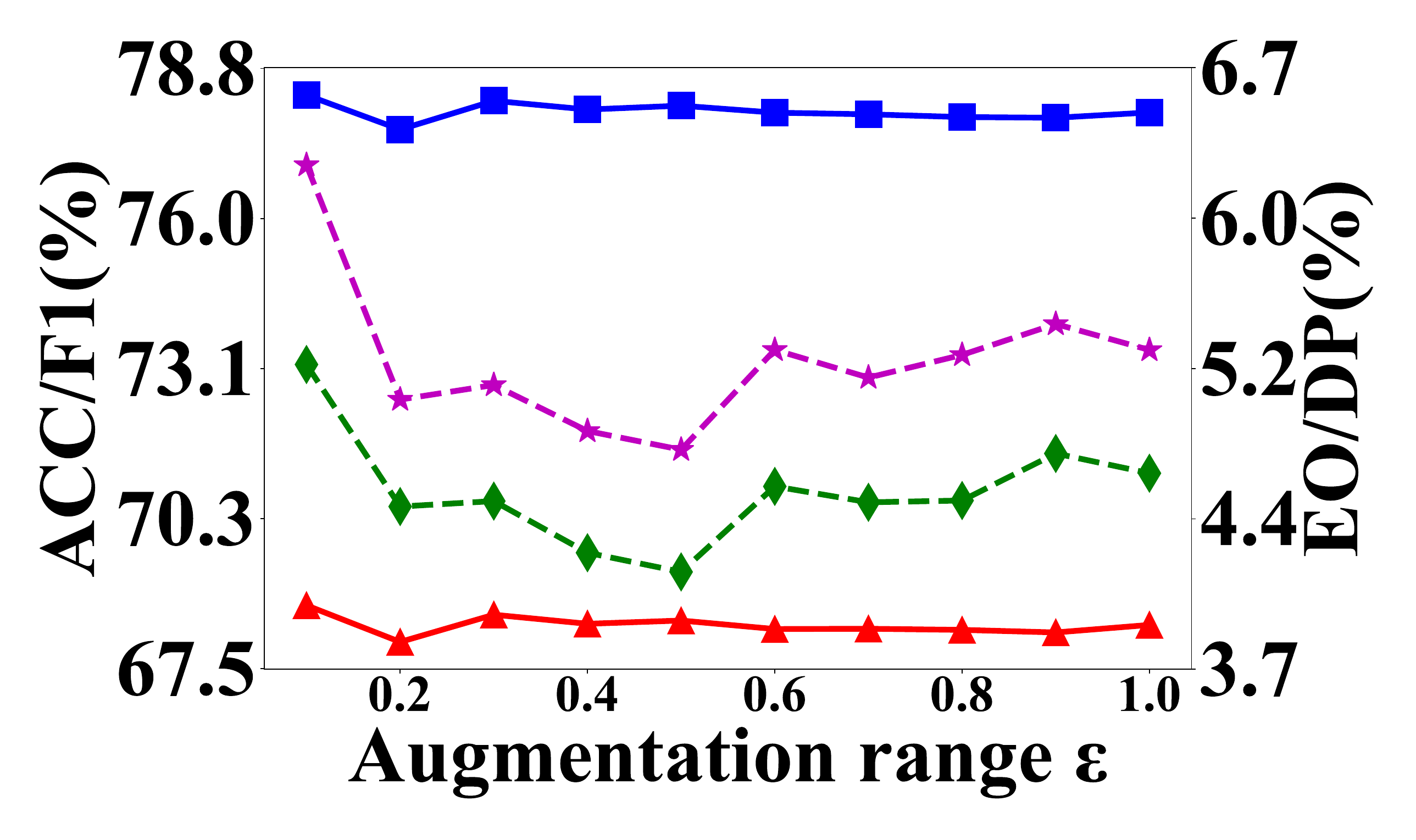}
        \caption{EdgePred-MinMax.}
    \end{subfigure}
    \begin{subfigure}[b]{0.23\linewidth}
        \includegraphics[width=\linewidth]{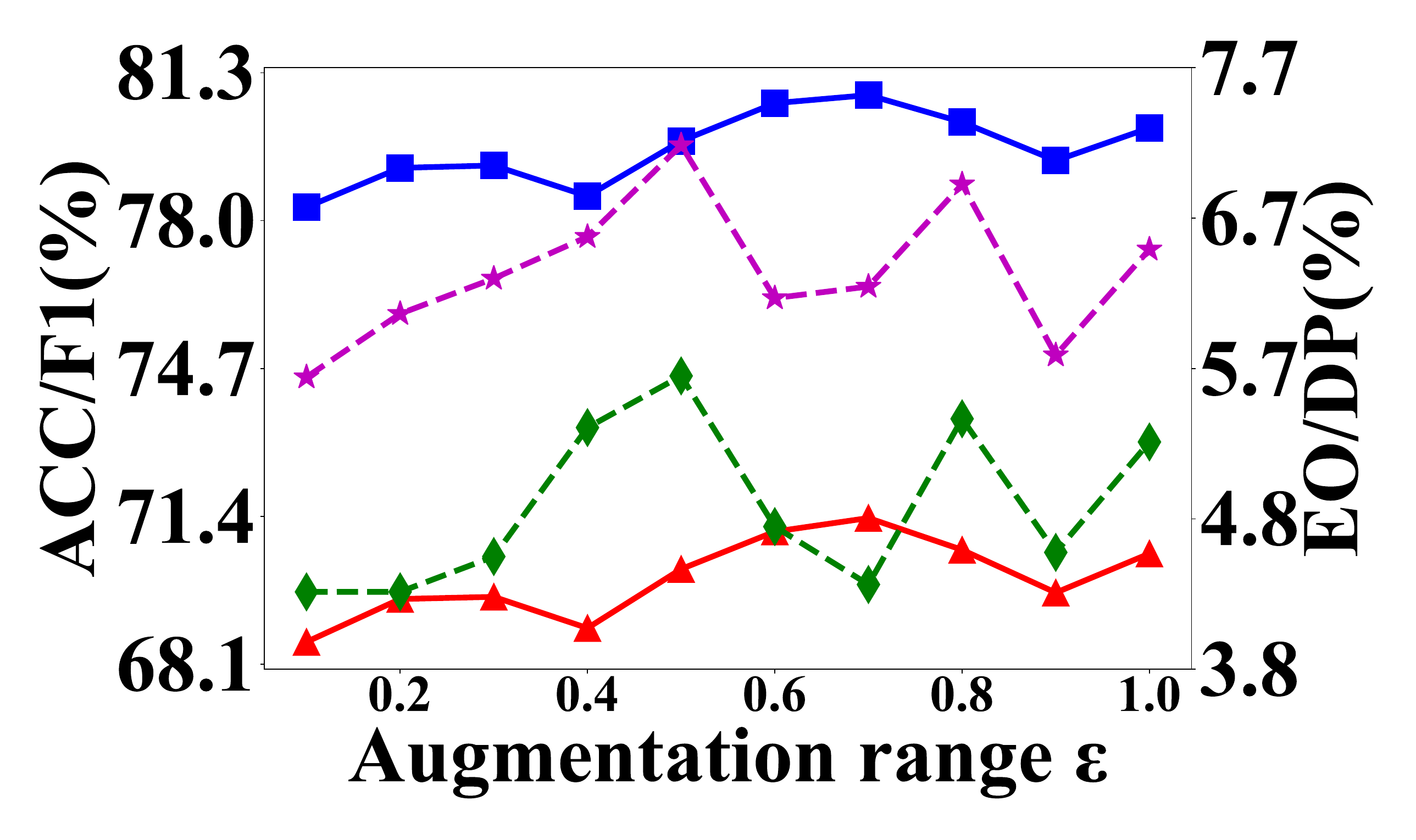}
        \caption{EdgePred-RandAT.}
    \end{subfigure}
    \caption{The effect of augmentation range $\epsilon$ to $\text{GraphPAR}_{minmax}$ and $\text{GraphPAR}_{RandAT}$ in the Credit dataset.}
    \label{fig:sen-cred}
\end{figure*}

\begin{figure*}[htbp]
    \begin{subfigure}[b]{0.23\linewidth}
        \includegraphics[width=\linewidth]{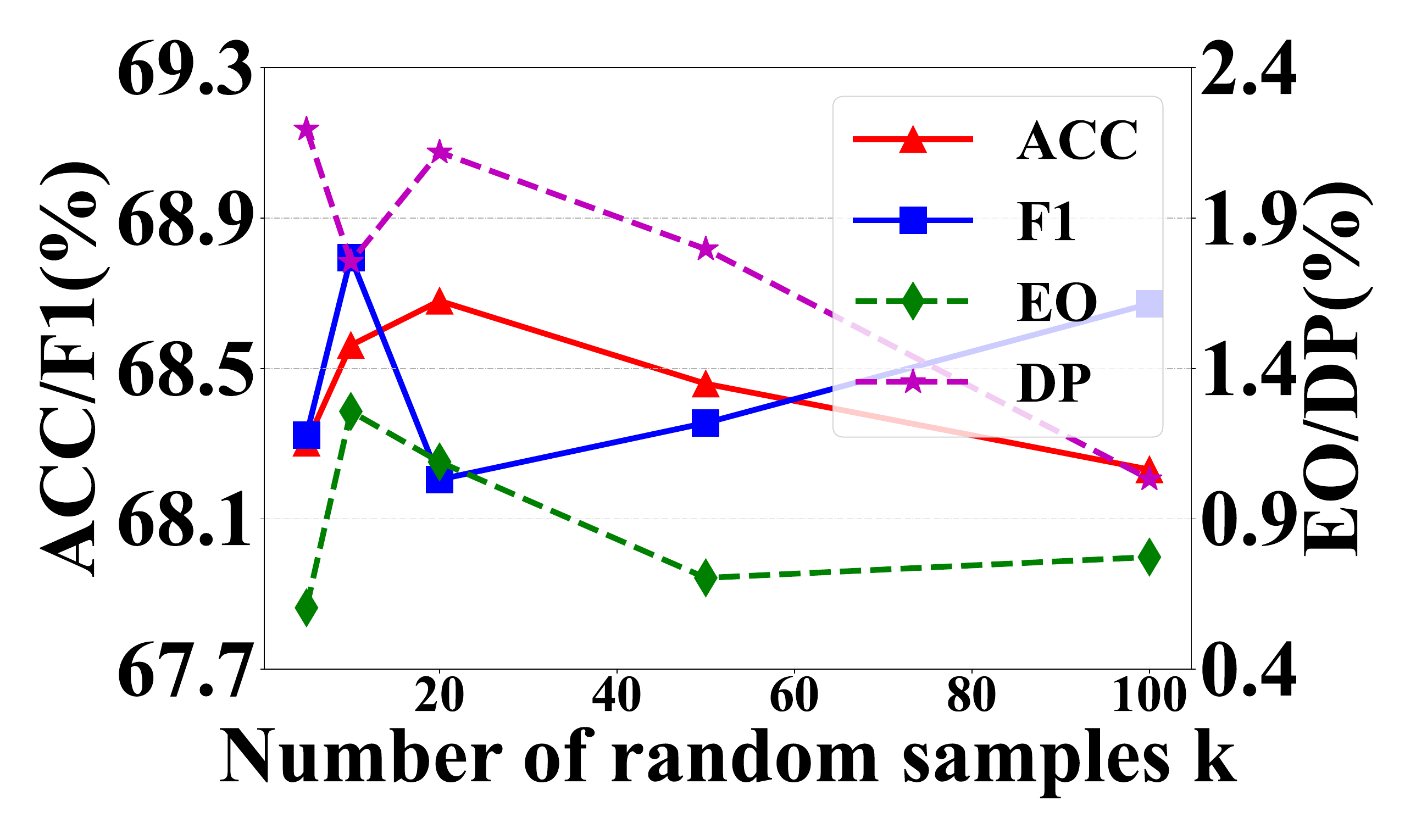}
        \caption{DGI-MinMax.}
    \end{subfigure}
    \begin{subfigure}[b]{0.23\linewidth}
        \includegraphics[width=\linewidth]{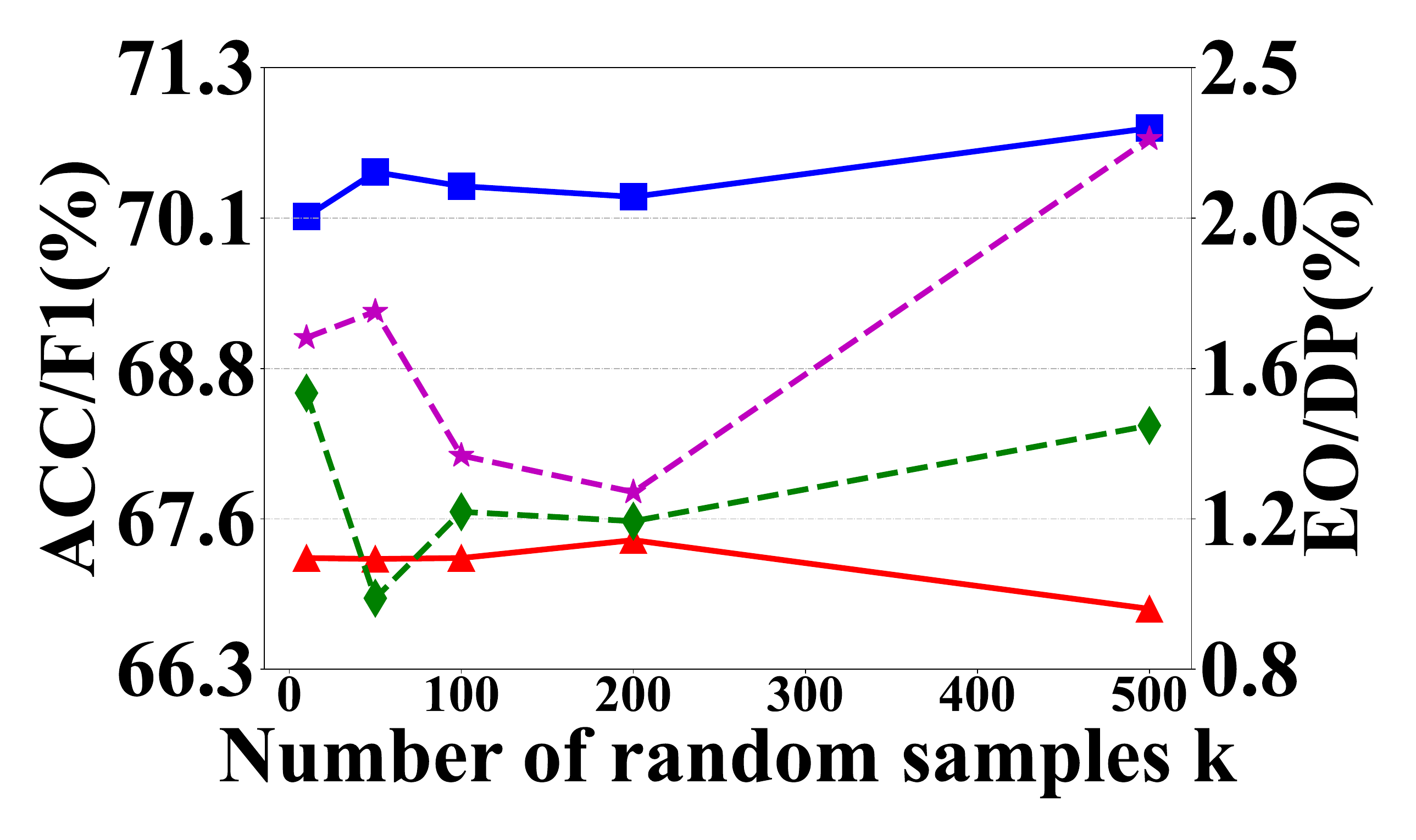}
        \caption{DGI-RandAT.}
    \end{subfigure}
    \begin{subfigure}[b]{0.23\linewidth}
        \includegraphics[width=\linewidth]{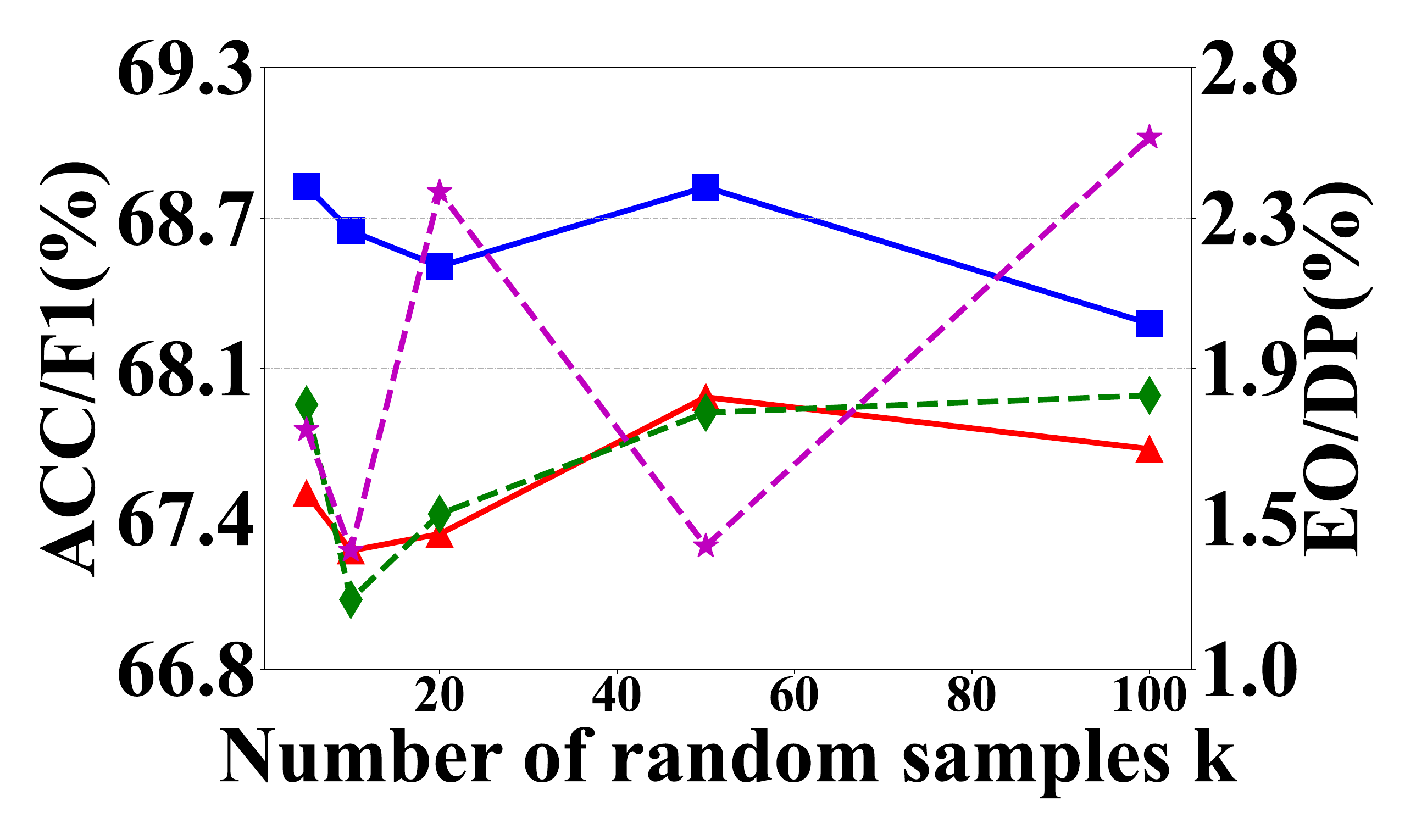}
        \caption{EdgePred-MinMax.}
    \end{subfigure}
    \begin{subfigure}[b]{0.23\linewidth}
        \includegraphics[width=\linewidth]{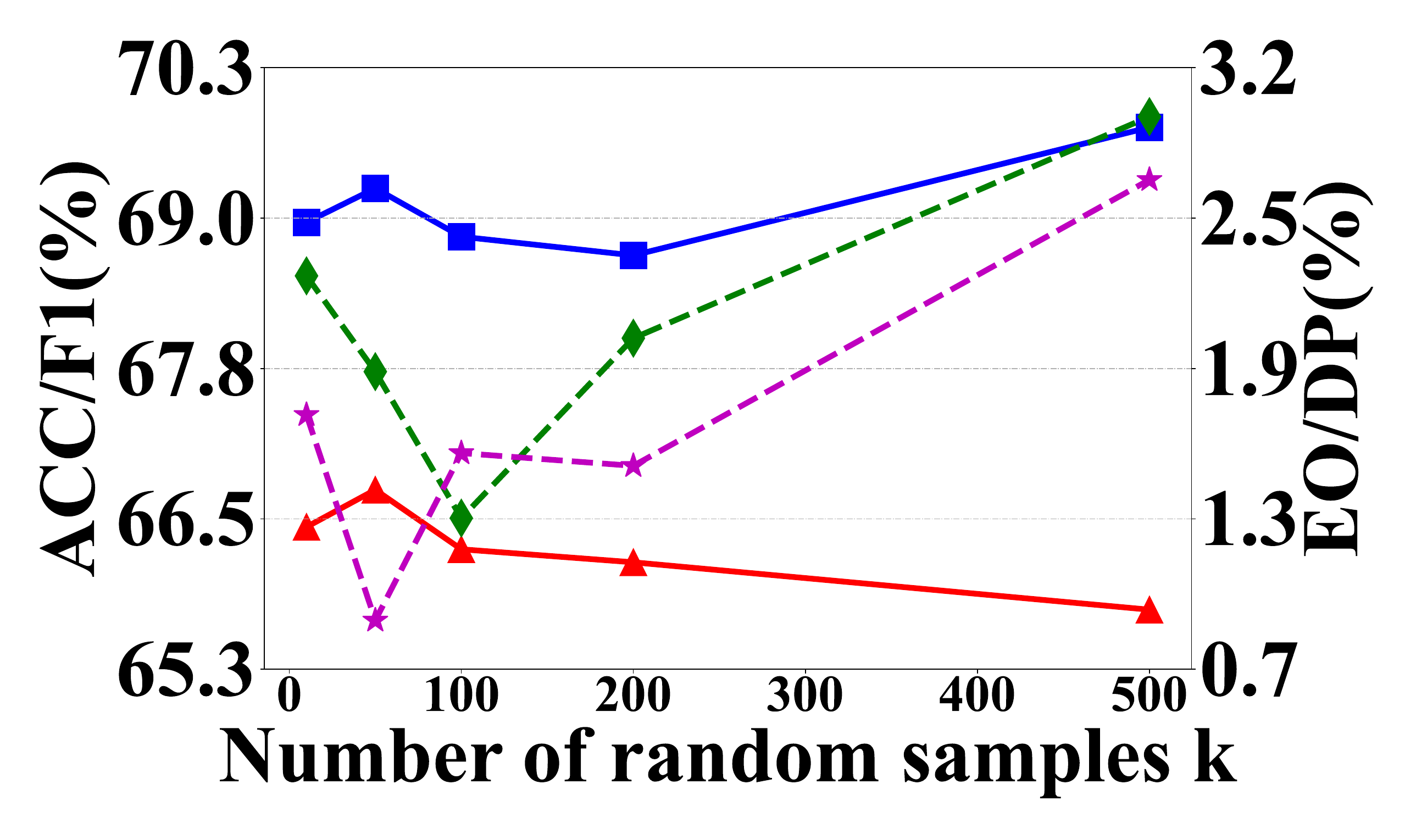}
        \caption{EdgePred-RandAT.}
    \end{subfigure}
    \caption{The effect of augmentation sample number $k$ to $\text{GraphPAR}_{minmax}$ and $\text{GraphPAR}_{RandAT}$ in the Pokec\_z dataset.}
    \label{fig:sen-pok-k}
\end{figure*}

\begin{figure*}[htbp]
    \begin{subfigure}[b]{0.23\linewidth}
        \includegraphics[width=\linewidth]{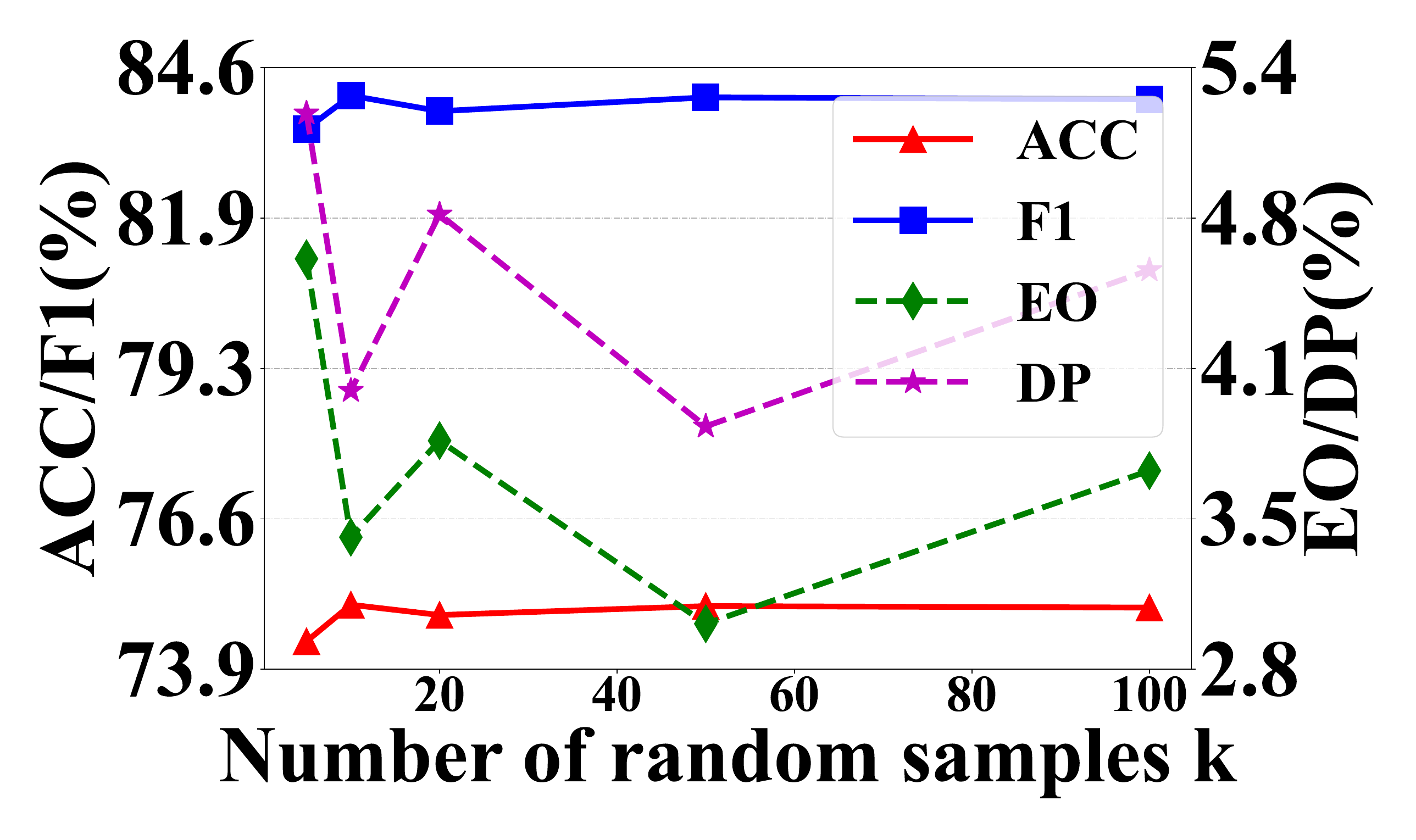}
        \caption{DGI-MinMax.}
    \end{subfigure}
    \begin{subfigure}[b]{0.23\linewidth}
        \includegraphics[width=\linewidth]{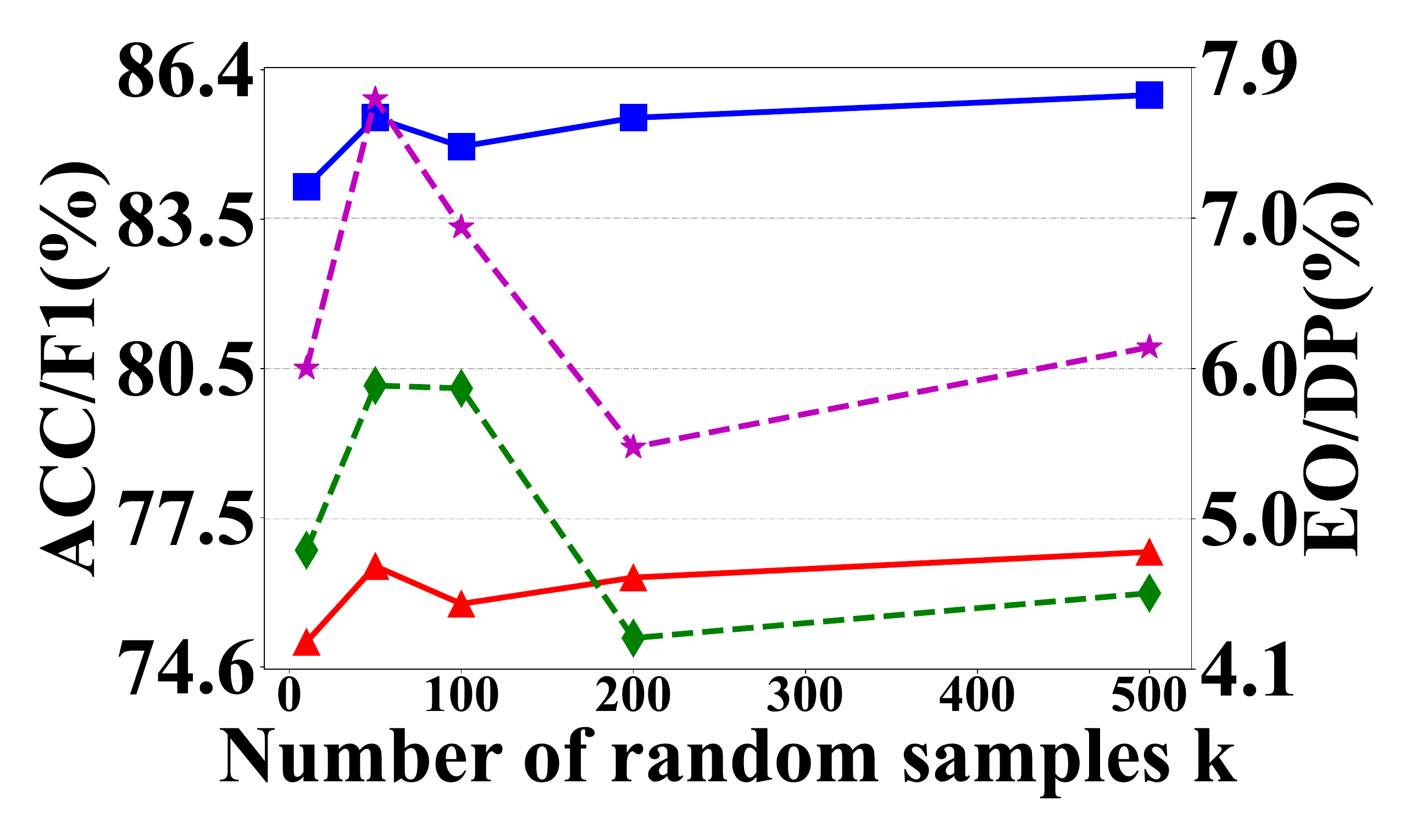}
        \caption{DGI-RandAT.}
    \end{subfigure}
    \begin{subfigure}[b]{0.23\linewidth}
        \includegraphics[width=\linewidth]{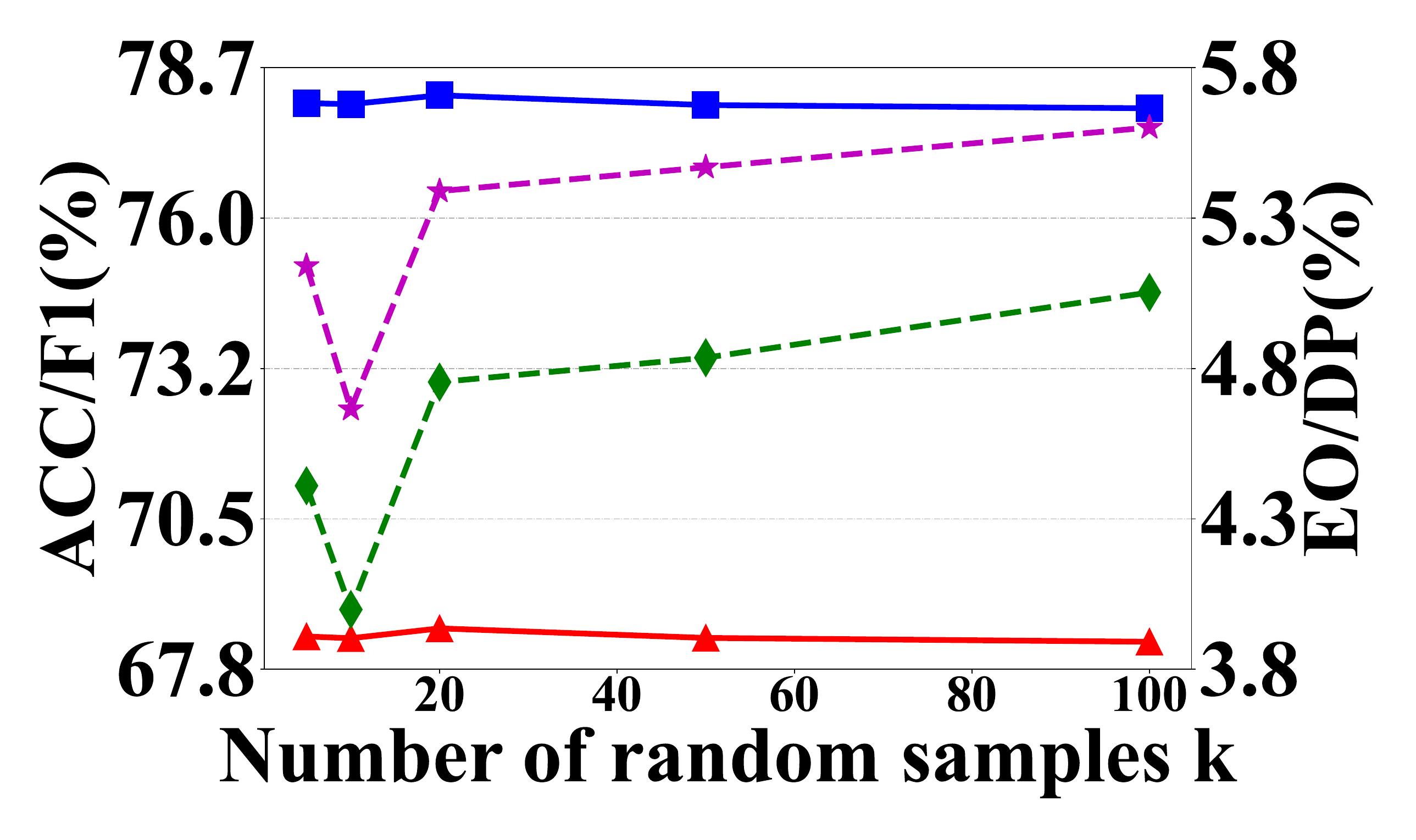}
        \caption{EdgePred-MinMax.}
    \end{subfigure}
    \begin{subfigure}[b]{0.23\linewidth}
        \includegraphics[width=\linewidth]{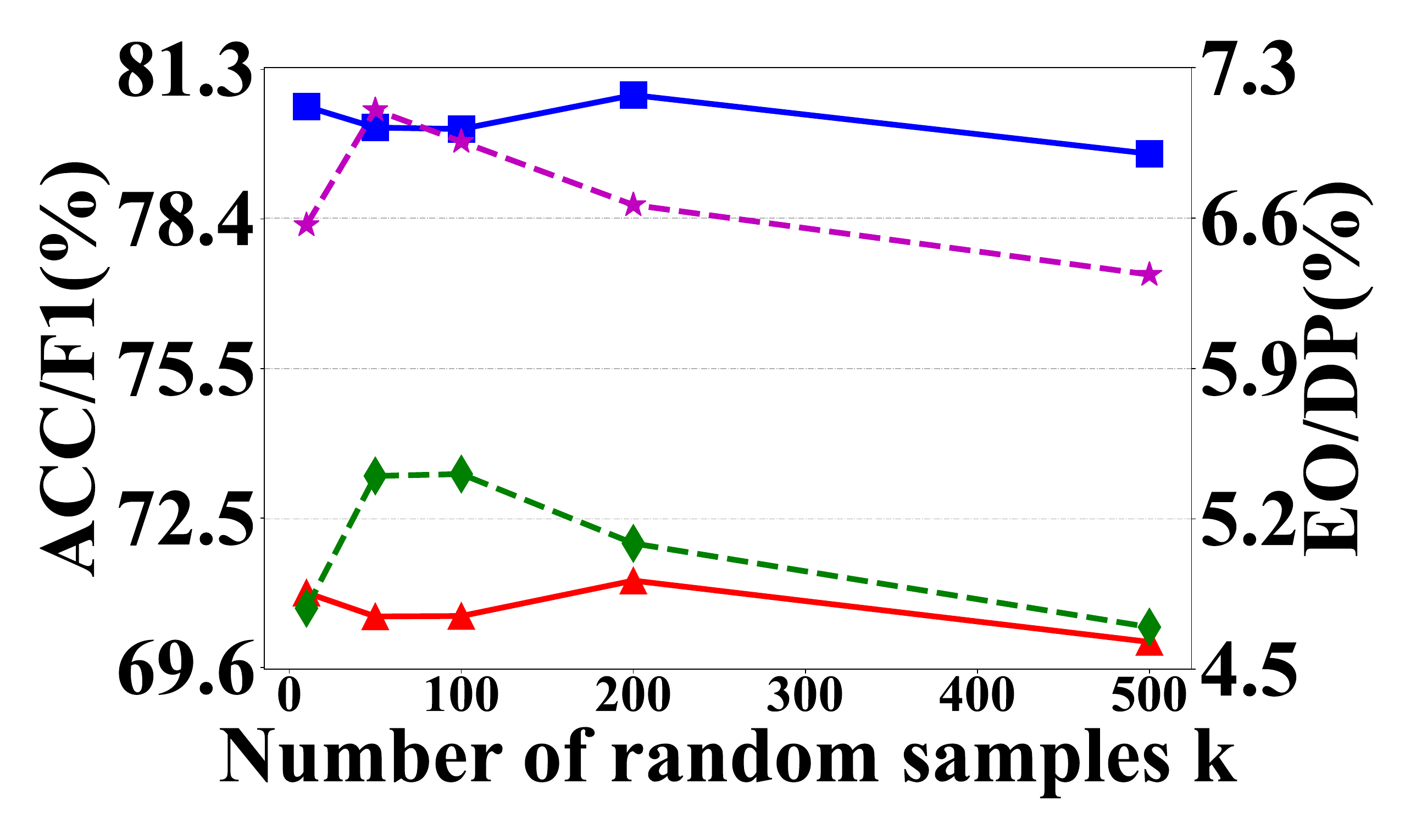}
        \caption{EdgePred-RandAT.}
    \end{subfigure}
    \caption{The effect of augmentation sample number $k$ to $\text{GraphPAR}_{minmax}$ and $\text{GraphPAR}_{RandAT}$ in the Credit dataset.}
    \label{fig:sen-cred-k}
\end{figure*}
We conduct a more detailed hyperparameter sensitivity analysis for GraphPAR, focusing on three key hyperparameters: the augmentation range $\epsilon$, the augmentation sample number $k$, and the fairness loss scale $\lambda$. They play a crucial role in shaping the prediction performance and fairness of GraphPAR, and understanding their sensitivity is vital for finding the best model for prediction performance and fairness.

\textbf{Augmentation range sensitivity ($\epsilon$).} The augmentation range $\epsilon$ dictates the range of linear interpolation on sensitive attribute semantics. Empirically, we find that the range of [0.2,0.4,0.8,1.0] works well for all datasets. An $\epsilon$ larger than 1 would probably harm the prediction accuracy. Within a certain range, the larger the augmentation range $\epsilon$, i.e., the larger the range of sensitive attributes considered, the model fairer. For example, as depicted in Figure ~\ref{fig:sen-cred} (a), when the PGM is DGI and the debiasing method is MinMax, the metrics of DP and EO tend to decrease with increasing  $\epsilon$ on the Credit dataset. 

\textbf{Fairness loss scale factor sensitivity ($\lambda$).} $\lambda$ is a scale factor for balancing accuracy and fairness. We find that different pre-training methods require different values of $\lambda$. As depicted in Figure ~\ref{fig:sen-lamda}, when the PGM is DGI, the optimal $\lambda$ is 0.7 in the Pokec\_z and Credit datasets. However, the optimal $\lambda$ is 0.2 when the PGM is EdgePred.

\textbf{Augmentation sample number sensitivity ($k$).} $k$ is the augmentation sample number for each node. According to Figure \ref{fig:sen-pok-k} and Figure \ref{fig:sen-cred-k}, the optimal $k$ is associated with the dataset, the pre-training method, and the adapter training strategy, but the general RandAT requires a larger $k$ value than MinMax.

\end{document}